%% file: mypaper.tex
\documentclass{article}

\usepackage{microtype}
\usepackage{graphicx}
\usepackage{subcaption}
\usepackage{booktabs} 
\usepackage{enumitem}

\usepackage{multicol}      
\usepackage{titletoc}     
\usepackage{placeins}    
\usepackage{xcolor}        

\definecolor{myred}{RGB}{220, 20, 60} 

\usepackage[accepted]{icml2026}

\usepackage{amsmath}
\usepackage{amssymb}
\usepackage{mathtools}
\usepackage{amsthm}
\usepackage{multirow}

\usepackage[capitalize,noabbrev]{cleveref}

\theoremstyle{plain}

\theoremstyle{definition}

\theoremstyle{remark}

\usepackage[title]{appendix}%

\usepackage[disable,textsize=tiny]{todonotes}

\icmltitlerunning{PACE: Parameter Change for Unsupervised Environment Design}

\begin{document}

\twocolumn[
  \icmltitle{PACE: Parameter Change for Unsupervised Environment Design}



  \begin{icmlauthorlist}
    \icmlauthor{Fang Yuan}{nudt,test,slk}
    \icmlauthor{Quanjun Yin}{nudt,slk}
    \icmlauthor{Siqi Shen}{fujian,xmu}
    \icmlauthor{Yuxiang Xie}{nudt}
    \icmlauthor{Junqiang Yang}{test,slk}
    \icmlauthor{Long Qin}{nudt,slk}    
    \icmlauthor{Junjie Zeng}{nudt,slk}
    \icmlauthor{Qinglun Li}{nudt,slk}
  \end{icmlauthorlist}

  \icmlaffiliation{nudt}{College of Systems Engineering, National University of Defense Technology, Changsha, China}
  \icmlaffiliation{test}{Test Center, National University of Defense Technology, Xi'an, China}
  \icmlaffiliation{slk}{State Key Laboratory of Digital Intelligent Modeling and Simulation, Changsha, China}
  \icmlaffiliation{fujian}{Fujian Key Laboratory of Urban Intelligent Sensing and Computing, Xiamen University, Xiamen, China}
  \icmlaffiliation{xmu}{Key Laboratory of Multimedia Trusted Perception and Efficient Computing, Ministry of Education of China, School of Informatics, Xiamen University, Xiamen, China}

  \icmlcorrespondingauthor{Junjie Zeng}{zengjunjie13@nudt.edu.cn}
  \icmlcorrespondingauthor{Qinglun Li}{liqinglun@nudt.edu.cn}

  \icmlkeywords{Machine Learning, ICML}

  \vskip 0.3in
]



\printAffiliationsAndNotice{}  

\begin{abstract}

Unsupervised Environment Design (UED) offers a promising paradigm for improving reinforcement learning generalization by adaptively shaping training environments, but it requires reliable environment evaluation to remain effective.
However, existing UED methods evaluate environments using indirect proxy signals such as regret, value-based errors, or Monte Carlo, which suffer from bias, high variance, or substantial computational overhead and fail to reflect agent realized learning progress.
To address these limitations, we propose \emph{Parameter Change Environment Design} (PACE), which evaluates an environment through the policy parameter change induced by training on that environment, directly grounding environment selection in realized learning progress.
Specifically, PACE assigns environment value using a first-order approximation of the policy optimization objective, where the improvement induced by an environment is proportional to the squared $\ell_2$ norm of the corresponding parameter update, enabling low-variance and computation-efficient evaluation without additional rollouts.
Experiments on MiniGrid and Craftax show that PACE consistently outperforms established UED baselines, achieving higher IQM and smaller Optimality Gap on OOD evaluations, including an IQM of $96.4\%$ and an Optimality Gap of $17.2\%$ on MiniGrid.

\end{abstract}

\begin{figure}[t]
  \centering
  \includegraphics[width=0.95\columnwidth]{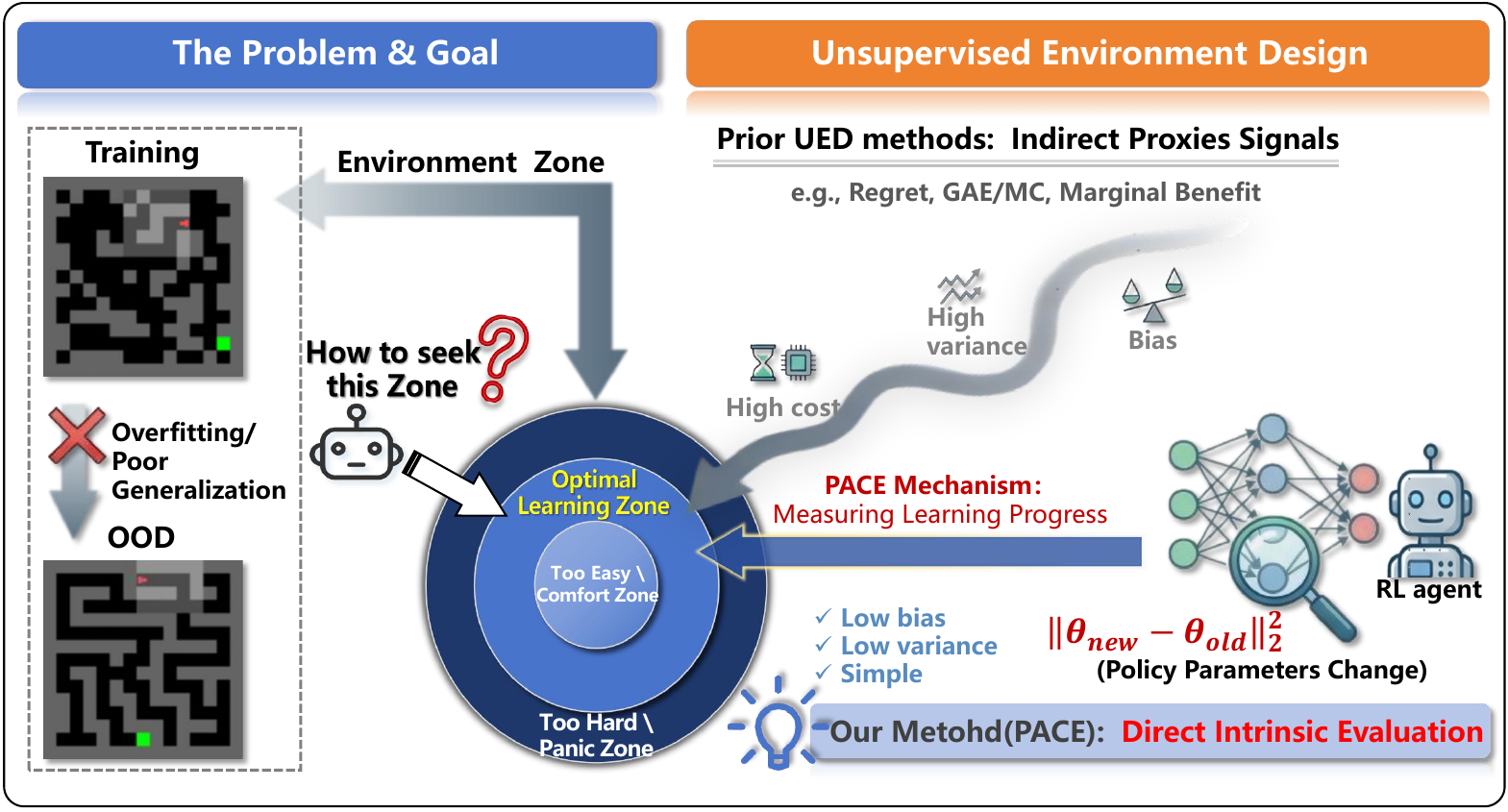}
  \caption{
Overview of Unsupervised Environment Design and the core idea of PACE.
Prior UED methods rely on indirect proxy signals (e.g., regret, GAE, or Monte Carlo return) to evaluate environments, which often suffer from bias, high variance, or high computational cost.
PACE directly measures agent learning progress via policy parameter change induced by environment interaction, providing a simple, low-variance, and intrinsic evaluation signal.
}
  \label{fig:F1}
\end{figure}

\input{chapter/1.introduction}

\input{chapter/2.Background}

\input{chapter/3.method}

\input{chapter/4.Experiments}

\input{chapter/7.Conclusion}

\bibliography{example_paper}
\bibliographystyle{icml2026}

\newpage
\appendix
\onecolumn
\input{chapter/8.appendix}


\end{document}

%% file: chapter/1.introduction.tex
\section{Introduction}

Reinforcement Learning (RL)~\cite{shakya2023reinforcement} achieves remarkable success in complex domains, including game playing~\cite{vinyals2019grandmaster,jain2024recent}, robotic control~\cite{elguea2023review,andrychowicz2020learning}, and many others~\cite{tang2025deep,towers2024gymnasium}. 
However, RL agents often suffer severe performance degradation when deployed in \emph{out-of-distribution} (OOD) environments, primarily due to overfitting to the training environment distribution.
This limitation highlights the central role of training environment design in reinforcement learning, and in particular the need to identify training environments that induce sustained learning progress as the agent improves.

Unsupervised Environment Design (UED)~\cite{dennis2020emergent} addresses this challenge by automatically shaping the training environment distribution, with the goal of inducing policies that continue to improve and generalize beyond observed environments.
Traditional approaches such as Domain Randomization (DR)~\cite{tobin2017domain} sample environment parameters uniformly at random, while minimax adversarial methods~\cite{pinto2017supervision} generate challenging environments through game-theoretic formulations.
In practice, both approaches often produce environments with highly inconsistent utility for learning, including instances that are either trivial or effectively unsolvable, which limits their ability to reliably support generalization, as these approaches do not explicitly account for whether a given environment induces effective learning progress at the agent current training stage. For example, As illustrated by the Panic Zone in Fig.~\ref{fig:F1}, the agent struggles to acquire knowledge in such environments.

UED reframes this problem by explicitly optimizing the training environment distribution.
Specifically, UED is defined as the task of using an underspecified environment to produce a distribution over fully specified environments that supports the continued learning of a given policy.
In this setting, training environments are generated by instantiating free parameters of an underspecified environment, and we refer to each such instantiated environment as a level.

A representative UED method, Protagonist Antagonist Induced Regret Environment Design (PAIRED)~\cite{dennis2020emergent}, evaluates environment value through \emph{regret}, defined as the performance gap between an agent and an optimal policy.
Because the optimal policy is generally unavailable, PAIRED relies on a teacher to generate environments and an expert to approximate optimal behavior.
Although effective, this formulation incurs substantial computational overhead due to repeated interactions among multiple agents and the environment, and it remains vulnerable to catastrophic forgetting during training~\cite{french1999catastrophic}.
To improve efficiency and stability, subsequent works—including Prioritized Level Replay (PLR)~\cite{jiang2021prioritized}, Robust Prioritized Level Replay (PLR$^{\perp}$)~\cite{jiang2021replay}, Adversarially Compounding Complexity by Editing Levels (ACCEL)~\cite{parker2022evolving}, and Marginal Benefit and Diversity driven Environment Design (MBeDED)~\cite{li2025marginal}—replace regret with alternative environment evaluation signals.
These methods employ proxy metrics such as Generalized Advantage Estimation (GAE), Monte Carlo (MC) returns, and rely on replay, environment editing, or mutation to construct the training distribution.

Despite substantial progress, existing UED methods still lack a reliable learning progress signal, and instead rely on environment evaluation criteria that are only indirectly related to agent realized improvement.
Regret-based criteria characterize theoretical learning potential rather than the improvement an agent actually achieves at its current training stage, which often leads to selecting levels beyond the agent learning capacity.
Replay-based methods that use GAE introduce estimation bias due to value function errors, while Monte Carlo--based criteria avoid this bias at the cost of high variance and substantial computational overhead.
As a result, current approaches struggle to provide an environment value signal that simultaneously reflects realized learning progress, maintains low variance, and remains computationally efficient.

We propose \textbf{Parameter Change Environment Design} (PACE), a simple and principled framework for environment evaluation in UED.
PACE evaluates environment value through policy parameter change induced by training on a level, directly grounding environment selection in agent learning progress.
This formulation avoids value estimation, return comparison, and additional environment interaction.

Our contributions are threefold:
\begin{itemize}[leftmargin=10pt, itemsep=1pt, topsep=1pt, parsep=0pt]
    \item We propose \emph{Parameter Change Environment Design} (PACE), a simple environment evaluation method for UED.
    PACE measures the value of a level through the magnitude of policy parameter change induced by training on that level, providing a direct and intrinsic signal of agent realized learning progress without value estimation, regret approximation, or additional environment interaction.

    \item We provide a theoretical analysis based on a first-order Taylor approximation of the policy optimization objective.
    Under a local gradient-aligned update, we show that the improvement induced by a level is proportional to the squared $\ell_2$ norm of the corresponding policy parameter update, establishing a principled connection between environment value and parameter change.

    \item We empirically evaluate PACE on MiniGrid and Craftax against established UED baselines.
    On MiniGrid OOD evaluation, PACE achieves an IQM of 0.964 and an Optimality Gap of 0.172, compared to 0.808 and 0.299 for the strongest baseline.
    On Craftax, PACE attains an IQM of 0.694 and an Optimality Gap of 0.341, outperforming the strongest baseline under both aggregate metrics.   
\end{itemize}

%% file: chapter/2.Background.tex
\section{Background}

In this section, we provide background on UED, including its problem formulation, generalization objective, and representative prior approaches. We defer a more comprehensive discussion of related work to Appendix~\ref{relatedwork}.

\subsection{Unsupervised Environment Design}

Unsupervised Environment Design (UED) studies how to train agents that generalize across a family of environments by actively shaping the distribution of training environments during learning. Following the formulation of~\cite{dennis2020emergent}, UED is formalized as an underspecified partially observable Markov decision process (UPOMDP) $\mathcal{M}=\langle \mathcal{A},\mathcal{O},\mathcal{S},\Phi,\mathcal{T},\mathcal{I},\mathcal{R},\gamma\rangle$, where $\Phi$ denotes a set of free environment parameters and $\mathcal{I}$ is observation function.

Fixing $\phi\in\Phi$ yields a fully specified POMDP $\mathcal{M}_\phi$, which we refer to as a \emph{level} $l$. For a policy $\pi$, its value in level $l$ (equivalently $\mathcal{M}_\phi$) is defined as
$V_\phi(\pi)=\mathbb{E}\!\left[\sum_{t=0}^{\infty}\gamma^t r_t\right]$.
Because $\Phi$ can be large or continuous, training on all possible levels is infeasible. As a result, the agent can only observe a small subset of levels during learning, which motivates UED methods that select or generate informative levels to support generalization.

\subsection{The Generalization Objective}

The core goal of UED is to train a policy that maximizes expected return on unseen environments.
Formally, we define the generalization objective as
\[
J(\theta) = \mathbb{E}_{\phi \sim P_{\text{test}}} \left[ V_{\phi}(\pi_{\theta}) \right],
\]
where $P_{\text{test}}$ denotes the distribution of unseen environments and
$\pi_{\theta}$ is the agent policy parameterized by $\theta$.

Since $P_{\text{test}}$ is unknown during training, we cannot directly optimize $J(\theta)$.
Instead, we optimize a surrogate objective based on a training distribution of environments,
denoted by $P_{\text{train}}$.
The training distribution is dynamically adjusted by the UED algorithm,
aiming to approximate the generalization objective.

To characterize how policy updates affect generalization, we define the policy gradient
\[
\Delta J(\theta)=\nabla_{\theta}J(\theta)=\mathbb{E}_{\phi\sim P_{\text{test}}}\!\left[\nabla_{\theta}V_{\phi}(\pi_{\theta})\right].
\]
UED algorithms therefore aim to select training environments whose induced updates align with the true generalization gradient $\Delta J(\theta)$.

\subsection{Existing Approaches to UED}

\paragraph{PAIRED.}
The first principled approach to UED is
\emph{Protagonist Antagonist Induced Regret Environment Design} (PAIRED).
Rather than minimizing the return of a single agent, PAIRED introduces an
unconstrained antagonist policy that cooperates with the environment generator.
The generator aims to maximize \emph{regret}, defined as the performance gap
between the antagonist and the protagonist in the same environment:
\[
\mathrm{REGRET}_{\phi}(\pi_P, \pi_A)
=
U_{\phi}(\pi_A) - U_{\phi}(\pi_P),
\]
where
\(
U_{\phi}(\pi) = \mathbb{E}_{\tau \sim \pi,\, l(\phi)}
\left[ \sum_{t=0}^{T} \gamma^t r_t \right]
\)
denotes the trajectory return of policy $\pi$.
Maximizing regret incentivizes the generator to produce environments that are
challenging yet solvable: the protagonist fails while the antagonist succeeds.
As the protagonist improves, the generator must construct increasingly complex
levels, which induces an adaptive curriculum aligned with the agent capability.

\paragraph{Prioritized Level Replay (PLR).}
To address the inefficiency of training an explicit antagonist,
Jiang et al.\ propose \emph{Prioritized Level Replay} (PLR).
PLR maintains a level buffer $\Lambda$ of size $K$ that stores previously visited
levels with high learning potential.
Instead of generating new environments, PLR samples levels from $\Lambda$ with
probability proportional to a trajectory-based score.

The score function measures the magnitude of the agent's prediction error along
a trajectory $\tau$ and is defined using the GAE:
\[
S_{\text{PLR}}(\tau)
=
\frac{1}{T}
\sum_{t=0}^{T}
\left|
\sum_{k=t}^{T} (\gamma \lambda)^{k-t} \delta_k
\right|,
\]
where
\(
\delta_k = r_k + \gamma V(s_{k+1}) - V(s_k)
\)
is the temporal-difference error.

\paragraph{Regret-Guided Replay: Positive Value Loss and MaxMC.}
To strengthen the theoretical connection between replay-based methods and regret,
Jiang et al.\ revisit PLR under a minimax regret perspective.
They introduce \emph{Positive Value Loss}, which only considers positive GAE values:
\[
S_{\text{PLR}}^{\perp}(\tau)
=
\frac{1}{T}
\sum_{t=0}^{T}
\max\!\left(
\sum_{k=t}^{T} (\gamma \lambda)^{k-t} \delta_k,
\, 0
\right).
\]

The same work further proposes Maximum Monte Carlo (MaxMC) as an alternative regret estimator. 
MaxMC replaces bootstrapped value targets with the highest return observed on the same level, using the score
$S_{\text{MaxMC}}(l)=R_{\max}(l)-V(s_0)$,
which reduces bias by decoupling the metric from current value estimates.

\vspace{-0.6em}
\paragraph{ACCEL.}
While PLR focuses on replaying informative past levels, ACCEL emphasizes
explicit difficulty control.
ACCEL samples environments from a random generator and applies targeted edits to
adjust difficulty in order to maintain a desired agent success rate.
This mechanism ensures that training levels remain near the agent performance
frontier while avoiding excessively hard or trivial environments.

\vspace{-0.6em}
\paragraph{Marginal Benefit.}
Li et al.~propose Marginal Benefit as an agent-centric criterion that directly measures realized improvement. Given a baseline policy $\pi_B$ and an updated policy $\pi_A$, the marginal benefit of level $l$ is defined as
$\mu_\phi(\pi_A,\pi_B)=V_\phi(\pi_A)-V_\phi(\pi_B)$.
By focusing on performance improvement rather than static difficulty, this metric aligns environment selection more closely with learning progress, but still requires additional rollouts and Monte Carlo estimation, which introduces nontrivial computational overhead and variance.

\begin{figure*}[ht!]
    \centering
    \includegraphics[width=0.85\textwidth]{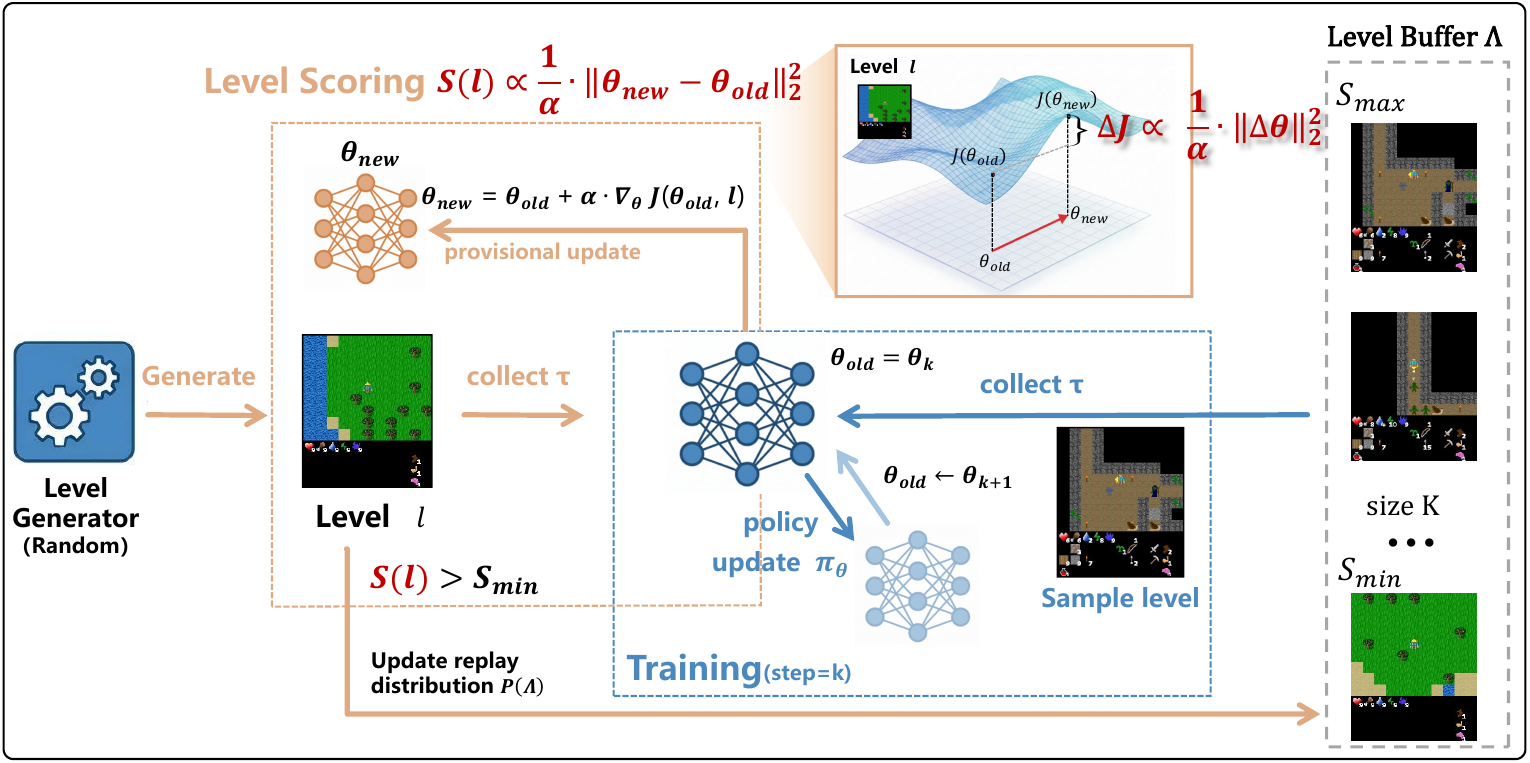}
    \caption{
    \textbf{Overview of the PACE framework.}
At training step $k$, the agent policy is $\pi_{\theta_k}$, and we set $\theta_{\text{old}} = \theta_k$.
The framework consists of two decoupled components: level scoring (orange) and policy training (blue).
In the level scoring stage, a level $l$ is generated by a random level generator, and a trajectory $\tau$ is collected using the current policy $\pi_{\theta_{\text{old}}}$.
A provisional policy update is then performed to obtain temporary parameters $\theta_{\text{new}}$, without writing back to the policy.
The 3D illustration depicts the objective landscape induced by level $l$, highlighting that improvement in the policy optimization objective correlates with the magnitude of the policy parameter change.
Motivated by the first-order analysis in Section~3.1, PACE uses policy parameter change as an intrinsic signal of realized learning progress, and uses this signal to construct the environment value of level $l$.
Specifically, PACE assigns a level score $S(l)$, defined in Eq.~(6), based on the magnitude of the induced parameter update.
If $S(l)$ exceeds the lowest score in the level buffer $\Lambda$, the new level replaces that entry.
In the training stage, levels are sampled from the level buffer $\Lambda$ according to a score-based replay distribution and used to update the policy, advancing the policy from $\pi_{\theta_k}$ to $\pi_{\theta_{k+1}}$.
The updated parameters are then assigned to $\theta_{\text{old}}$ for subsequent level scoring.
Overall, PACE iteratively leverages policy parameter change as a learning progress signal to select training levels that match the agent current learning capacity.
    }
    \label{fig:F2}
\end{figure*}

While existing methods advance Unsupervised Environment Design from different perspectives, their environment evaluation metrics still suffer from fundamental limitations.
PAIRED maximizes regret, which reflects the theoretical learning potential of an environment rather than the agent realized capability at its current training stage, and therefore tends to select levels that exceed the agent learning capacity and yield limited practical improvement.
Replay-based methods such as PLR and ACCEL show strong empirical performance under random environment generation, but their evaluation metrics remain imperfect.
Specifically, these methods rely on Generalized Advantage Estimation, whose bias depends on value function errors and bias--variance trade-off parameters, potentially distorting environment value assessment.
Monte Carlo--based metrics remove this bias but introduce high variance and substantial computational cost, while Marginal Benefit further amplifies both variance and overhead by comparing MC from different policy snapshots.

Overall, existing UED methods either rely on proxy signals weakly correlated with realized learning progress, or incur substantial estimation variance and computational cost.
This motivates the need for a simple and intrinsic environment value signal that directly reflects agent learning progress.

%% file: chapter/3.method.tex
\section{Approach}
\label{sec:method}

In this section, we introduce \textbf{Parameter Change Environment Design (PACE)}, a UED method that uses policy parameter updates as a simple and precise proxy for agent learning progress.
In Section~\ref{sec:pace_value}, we formalize environment value from the perspective of performance improvement and provide the corresponding theoretical derivation.
Section~\ref{sec:pace_algorithm} then describes how to select and reuse high-value environments in practice.

\subsection{Parameter Change as a Signal of Learning Progress}
\label{sec:pace_value}

Let $\pi_{\theta_{\text{old}}}$ denote the current policy before interacting with a level $l$.
After collecting trajectories $\tau$ in $l$ and applying a single local policy update, the policy parameters change from $\theta_{\text{old}}$ to $\theta_{\text{new}}$, yielding a new policy $\pi_{\theta_{\text{new}}}$.
We characterize the learning progress induced by level $l$ through the improvement in the policy optimization objective, and use this quantity to define the environment value in UED:
\begin{equation}
    \Delta J(l) = J(\theta_{\text{new}}, l) - J(\theta_{\text{old}}, l).
\end{equation}

In reinforcement learning, evaluating $J(\theta, l)$ relies on stochastic return estimation.
Directly estimating $\Delta J(l)$ therefore requires additional environment interactions and introduces substantial estimation noise.
We instead seek a deterministic and intrinsic approximation of $\Delta J(l)$ that depends only on the policy update induced by the level.

We derive such an approximation under an idealized local analysis that considers a single gradient-based update.
For a fixed level $l$, consider the objective $J(\theta, l)$ and apply a first-order Taylor expansion around the current parameters $\theta_{\text{old}}$:
\begin{equation}
    J(\theta_{\text{new}}, l) \approx J(\theta_{\text{old}}, l)
    + \nabla_{\theta} J(\theta_{\text{old}}, l)^\top (\theta_{\text{new}} - \theta_{\text{old}}).
\end{equation}
Assume a single-step update that follows the local gradient of the optimization objective,
\begin{equation}
    \theta_{\text{new}} = \theta_{\text{old}} + \alpha \nabla_{\theta} J(\theta_{\text{old}}, l),
\end{equation}
where $\alpha$ denotes the step size.
Rearranging yields
\begin{equation}
    \nabla_{\theta} J(\theta_{\text{old}}, l)
    = \frac{1}{\alpha} (\theta_{\text{new}} - \theta_{\text{old}}).
\end{equation}
Substituting into the Taylor expansion gives a first-order approximation of the objective improvement:
\begin{equation}
\begin{aligned}
    \Delta J(l)
    &= J(\theta_{\text{new}}, l) - J(\theta_{\text{old}}, l) \\
    &\approx \nabla_{\theta} J(\theta_{\text{old}}, l)^\top (\theta_{\text{new}} - \theta_{\text{old}}) \\
    &= \frac{1}{\alpha} (\theta_{\text{new}} - \theta_{\text{old}})^\top (\theta_{\text{new}} - \theta_{\text{old}}) \\
    &= \frac{1}{\alpha} \| \theta_{\text{new}} - \theta_{\text{old}} \|_2^2.
\end{aligned}
\end{equation}
This analysis highlights a key geometric relationship: under a local gradient-aligned update, the improvement in the optimization objective induced by a level is proportional to the squared $\ell_2$ norm of the resulting policy parameter change, up to a constant factor.
We emphasize that this derivation relies on idealized assumptions and serves as theoretical motivation rather than a precise characterization of the optimization dynamics used in practice.
It motivates the use of policy parameter change as a simple and intrinsic proxy for environment value, which we operationalize in the following section.

\subsection{PACE Algorithm}
\label{sec:pace_algorithm}

Based on the above analysis, we propose Parameter Change Environment Design (PACE), a UED framework that uses policy parameter change as an intrinsic signal of realized learning progress, which is then used to construct environment value for level selection. An overview of the framework is illustrated in Figure~\ref{fig:F2}.

First, PACE evaluates levels by collecting a trajectory $\tau$ on a newly generated level $l$ and performing a provisional policy update for scoring purposes.
In this framework, we adopt a simple random distribution as the level generator, which provides a diverse set of environments while remaining conceptually and computationally lightweight.

We then compute the level score according to Eq.~\ref{Sl}, which measures the magnitude of the induced policy parameter change. 
Based on this score, the level is inserted into the level buffer $\Lambda$ by replacing the lowest-scoring level when the buffer is full.
As a result, this procedure ensures that $\Lambda$ consistently retains levels that induce substantial policy parameter updates.
The complete procedure is summarized in Algorithm~1.
\begin{equation}
\label{Sl}
    S(l) = \frac{1}{\alpha} \| \theta_{\text{new}} - \theta_{\text{old}} \|_2^2 .
\end{equation}
In addition, as illustrated in Figure~\ref{fig:F2}, PACE samples levels from the level buffer $\Lambda$ during training using a score-based prioritized distribution.
Specifically, we adopt rank-based prioritization, where levels are ranked by their scores and sampled according to
\begin{equation}
\label{Pl}
    P(l_i) = 
    \frac{\left( \frac{1}{\text{rank}(S(l_i))} \right)^{1/\beta}}
    {\sum_{l_j \in \Lambda} \left( \frac{1}{\text{rank}(S(l_j))} \right)^{1/\beta}},
\end{equation}
where $\beta$ is a temperature parameter that controls the extent to which the inverse rank $\frac{1}{\text{rank}(S(l_i))}$ influences the resulting sampling distribution.

At the same time, as training progresses, the replay distribution may gradually concentrate on a subset of frequently sampled levels.
Levels that are not revisited for extended periods risk being neglected, even if they were previously informative.
Such imbalance can lead to catastrophic forgetting, where the agent loses proficiency on under-sampled levels.
To mitigate this effect, as shown in Figure~2, we also incorporate a staleness-aware mechanism that encourages periodic re-evaluation of stored levels.
By promoting the replay of levels that have not been sampled recently, this mechanism helps maintain coverage of the level buffer and stabilizes training over time.
For clarity, the prioritized sampling distribution and the staleness-aware mechanism are omitted from Algorithm~\ref{alg:pace}.

\paragraph{Strength of PACE.}
PACE offers several advantages over existing UED approaches by grounding environment evaluation in the dynamics of policy improvement.
First, the proposed score $S(l)$ provides a low-variance signal for environment value.
Unlike regret- or return-based criteria that rely on Monte Carlo estimation or value prediction, PACE evaluates a level solely through the induced policy parameter update, avoiding additional rollout-based evaluation and eliminating compounded return variance.

Second, PACE aligns environment selection with realized learning progress.
By measuring how much a level contributes to advancing the policy parameters, the evaluation signal directly reflects improvement in the optimization objective, rather than theoretical difficulty or prediction error.
This intrinsic coupling between environment value and policy updates encourages the selection of levels that match the agent current learning capacity, which is critical for robust generalization.

Third, PACE is computationally efficient.
Computing $S(l)$ requires only in-memory parameter differencing and incurs negligible overhead beyond a standard policy update.
In contrast, methods such as Marginal Benefit require repeated policy evaluation on the same environment, leading to substantially higher interaction cost.
As a result, PACE scales naturally to large environments and long training horizons, while remaining simple to implement within standard reinforcement learning pipelines.

\begin{algorithm}[tb]
  \caption{PACE}
  \label{alg:pace}
  \begin{algorithmic}
    \STATE {\bfseries Input:} Level buffer size $K$, initial fill ratio $\rho$, replay probability $p$, replay distribution $P$, level generator
    \STATE {\bfseries Initialize:} Policy $\pi_{\theta_{\text{old}}}$, level buffer $\Lambda$
    \STATE Sample $K \cdot \rho$ initial levels to populate $\Lambda$
    \WHILE{training not converged}
      \STATE Sample replay decision $\epsilon \sim \mathcal{U}[0,1]$
      \IF{$\epsilon \ge p$}
        \STATE Sample a new level $l$ from the level generator
      \ELSE
        \STATE Sample level $l$ from $\Lambda$ according to $P$
        \STATE Update policy to $\pi_{\theta}$ on level $l$
      \ENDIF
      \STATE Collect trajectory $\tau$ on level $l$
      \STATE Provisional update to obtain $\theta_{\text{new}}$ for scoring
      \STATE Compute level score $S(l)$ (Eq.~\ref{Sl})
      \IF{$S(l)$ exceeds the lowest score in $\Lambda$}
        \STATE Update $\Lambda$ with level $l$
      \ENDIF
      \STATE Set $\theta_{\text{old}} \leftarrow \theta$
    \ENDWHILE
  \end{algorithmic}
\end{algorithm}

%% file: chapter/4.Experiments.tex
\section{Experimental results}

In this section, we present experimental results on the \emph{MiniGrid} and \emph{Craftax} benchmarks to demonstrate zero-shot generalization when a trained agent transfers to OOD environments.
We compare \textbf{PACE} against several representative UED baselines, including DR~\cite{tobin2017domain}, PLR~\cite{jiang2021prioritized}, PLR\(^{\perp}\)~\cite{jiang2021replay}, ACCEL~\cite{parker2022evolving}. PAIRED and MBeDED rely on environment generation through an RL teacher, which introduces additional policy instances and environment interactions and leads to a substantially different computational model. Therefore, we leave a systematic comparison with these methods to future work. Nevertheless, we present a detailed analysis of computational complexity and evaluation stability in Appendix~\ref{ComplexityAnalysis}.
In all cases, we train a student agent using Proximal Policy Optimization (PPO)~\cite{schulman2017proximal} with the same policy architecture and optimizer. A complete list of hyperparameters for each experiment is provided in Table~\ref{tab:hyperparameters} in Appendix~\ref{Results}.
We show all performance metrics as a function of the number of gradient updates applied to the student policy. The corresponding total number of environment interactions is reported in Table~\ref{tab:env_steps} in Appendix~\ref{Results}.
We implement all experiments using \texttt{JaxUED}~\cite{coward2024jaxued}, a publicly available JAX-based framework for UED.
MiniGrid experiments run on a single RTX 5090 GPU (32G), while Craftax experiments run on a single A100 GPU(80G).

We begin with a partially observable navigation environment and evaluate transfer performance on human-designed levels in MiniGrid~\cite{chevalier2018minimalistic}. 
Then, We compare performance in Craftax~\cite{matthews2024craftax}, a JAX-based benchmark for open-ended reinforcement learning with substantially more complex dynamics, including multi-floor worlds, diverse enemies, and long-horizon objectives. 
Craftax builds upon the Crafter environment~\cite{hafner2021benchmarking} and extends it with mechanics inspired by Roguelike games and NetHack~\cite{kuttler2020nethack}.

\subsection{MiniGrid Environments}

The MiniGrid environment is widely used in UED research. 
It supports rapid construction of diverse levels by adjusting parameters such as the number of obstacles and maze size, which enables flexible control over task difficulty. 
Although the overall task structure is relatively simple, training robust agents typically requires large-scale training. 
Therefore, in this experiment, we train the student agent for 30k policy updates, which corresponds to 245,760,000 environment interaction steps.
We use an LSTM-based actor--critic policy with partially observable local observations as input. 
During training, all training levels use mazes with a size of $15\times15$ tiles. For all DR-based methods, we generate training levels by performing 100 random obstacle placement operations.
Each placement samples uniformly from grid locations and allows sampling positions that already contain obstacles. 
As a result, the number of obstacles in the final level is a random variable with an expected value of approximately 74.
For ACCEL, we begin with empty rooms and randomly edit block locations by adding or removing blocks, as well as modifying the goal location.

We follow definitions from prior work when selecting environment evaluation metrics. 
Specifically, PLR uses the absolute value of the GAE as metric, following~\cite{jiang2021prioritized}. 
PLR\(^{\perp}\) and ACCEL adopt MaxMC as metric, following~\cite{jiang2021replay, parker2022evolving}.
We evaluate zero-shot transfer performance on 12 human-designed levels, with a maximum episode length of 250 steps. The detailed levels can be found in Figure~\ref{fig:minigrid_12_levels} in Appendix~\ref{Results}.
All methods use 10 random seeds for training, and we evaluate the policy obtained at the end of PPO training. 
All experimental settings match those in prior work.

\begin{figure}[t]
  \centering
  \includegraphics[width=0.85\columnwidth]{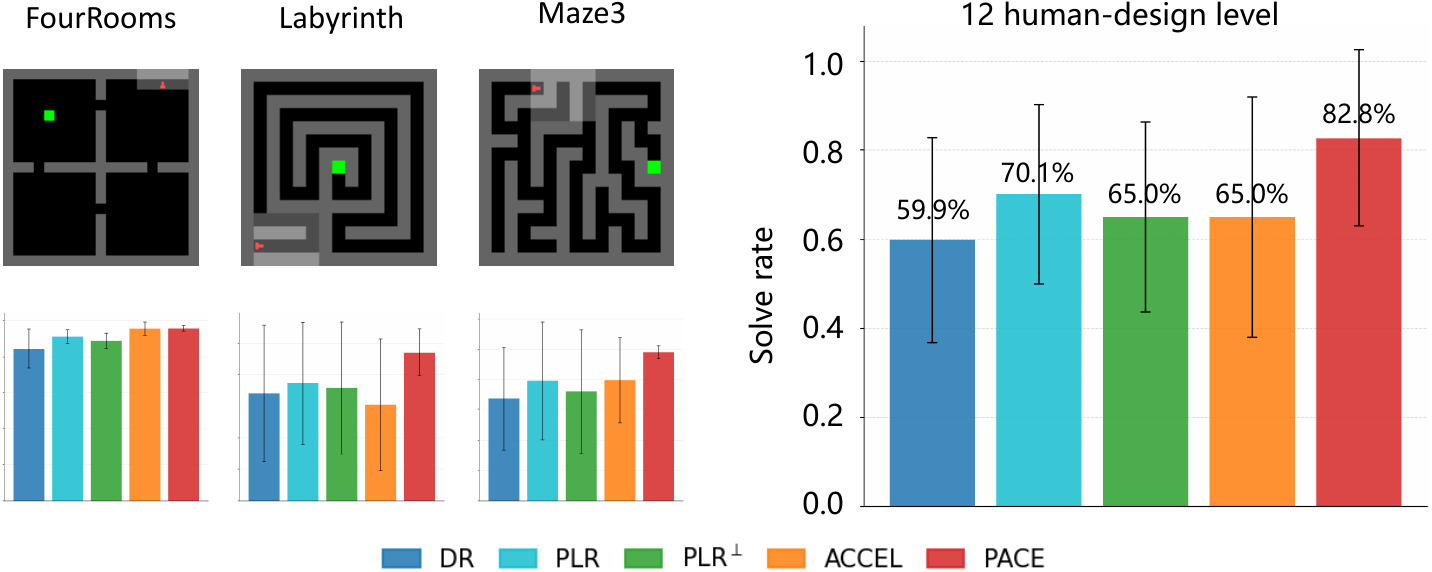}
  \caption{Zero-shot transfer performance on human-designed MiniGrid levels. Plots show the mean solve rate with one standard deviation on representative OOD levels over 10 runs. Moreover, PACE demonstrates smaller standard deviation than baselines, suggesting improved stability in zero-shot transfer performance.}
  \label{fig:minigrid}
\end{figure}

In Figure~\ref{fig:minigrid}, we select three representative OOD test levels with progressively increasing difficulty and report the mean and variance of solve rates for each method on these levels. 
We observe that on the relatively easy FourRooms level, all methods achieve solve rates above 84\%, indicating that this class of simple OOD environments poses little challenge for most approaches. 
However, as level complexity increases, performance differences among methods become increasingly pronounced.
On the more challenging Labyrinth and StandardMaze3 levels, our method consistently achieves substantially higher and more stable solve rates than the baseline methods. 
In particular, our method maintains solve rates above 90\% with significantly smaller variance, demonstrating stronger robustness in complex OOD environments.

In addition, the right panel of Figure~\ref{fig:minigrid} summarizes the average solve rate and variance across 12 human-designed test levels. 
Overall, our method improves the average solve rate by more than 12\% compared with the baselines and exhibits the smallest variance among all methods. 
These results indicate that our approach achieves more stable and consistent generalization across OOD levels with varying degrees of complexity. 
Detailed complete evaluation results appear in Figure~\ref{fig:mazesolve} and Table~\ref{tab:full_minigrid_results} in Appendix~\ref{Results}.

\begin{figure}[t]
  \centering
  \includegraphics[width=0.85\columnwidth]{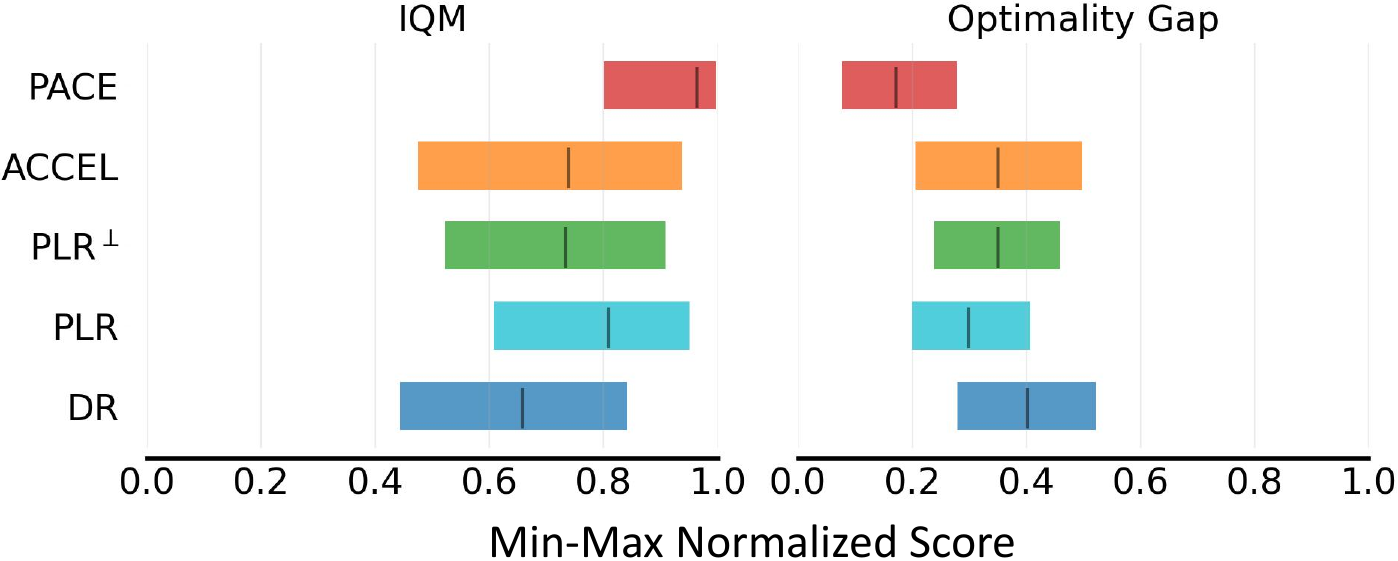}
  \caption{Aggregate OOD test performance on maze environments. Overall, PACE consistently dominates the performance–robustness trade-off, achieving both higher central performance and tighter performance dispersion across OOD maze environments.}
  \label{fig:IQMmaze}
\end{figure}

To evaluate generalization across OOD levels with varying complexity from a distributional perspective, we adopt the Interquartile Mean (IQM) and Optimality Gap metrics from the \texttt{rliable}~\cite{agarwal2021deep} library to summarize the distribution of solved rates across methods.
Figure~\ref{fig:IQMmaze} reports the corresponding results.
PACE achieves an IQM of 96.4\% in terms of solved rates, which is substantially higher than all baseline methods.
The strongest baseline, PLR, attains an IQM of 80.8\%.
Inspection of the IQM confidence intervals shows that PACE consistently maintains stronger performance in the central region of the solved-rate distribution across different OOD levels.
Under the Optimality Gap metric, PACE also achieves the smallest gap, with a value of 17.2\%, which is notably lower than the baselines.
Complete numerical results and statistical significance tests appear in Table~\ref{tab:iqm_gap_results} in Appendix~\ref{Results}.

\begin{figure}[t]
  \centering
  \includegraphics[width=0.85\columnwidth]{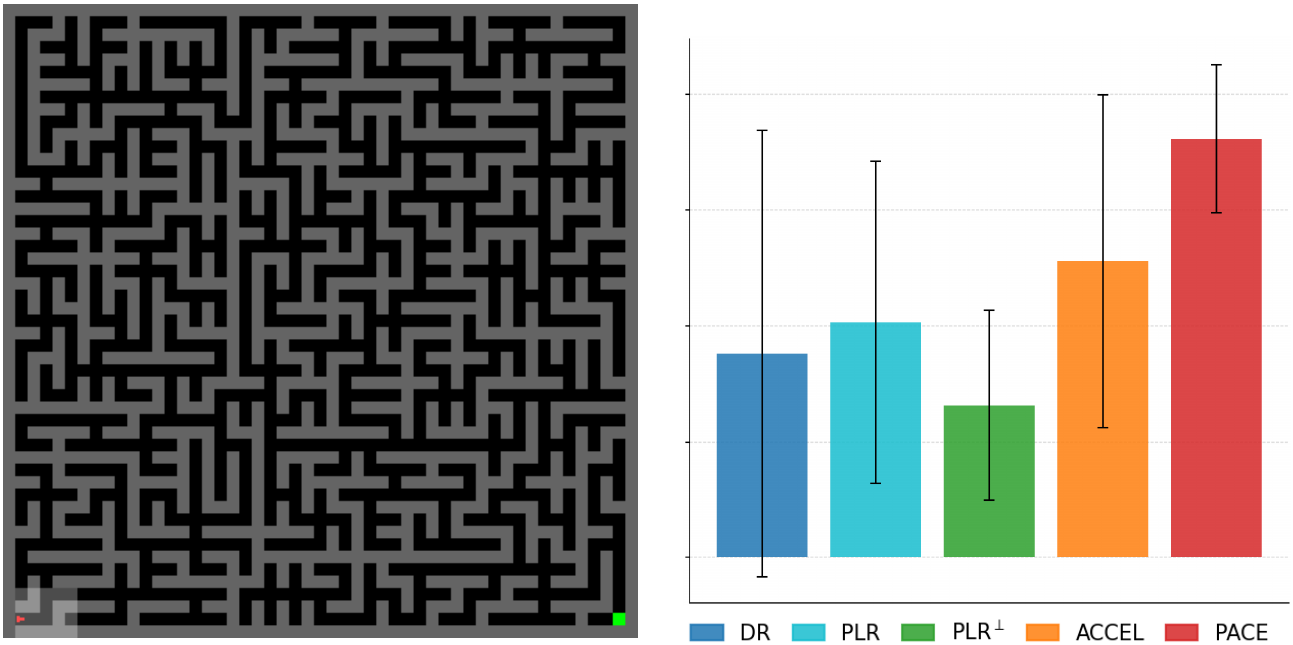}
  \caption{Zero-shot generalization to a large-scale procedurally generated maze. PACE attains a 72.3\% solve rate, substantially outperforming all baselines and highlighting superior generalization to large-scale OOD mazes. Bars show the mean and standard deviation over 10 runs.}
  \label{fig:PerfectMazeLarge}
\end{figure}

Finally, we evaluate zero-shot transfer performance on the largest version of PerfectMaze, a procedurally generated maze environment, as shown in Figure~\ref{fig:PerfectMazeLarge}. 
This setting contains levels with 51$\times$51 tiles and a maximum episode length of 3k steps, which is substantially larger than the environments used during training.
We evaluate agents for 100 episodes per training seed, using the same policy checkpoints as in Figure~\ref{fig:minigrid}. 
As shown in Figure~\ref{fig:PerfectMazeLarge}, PACE significantly outperforms all baseline, achieving a solve rate of 72.3\%.
These results further demonstrate that PACE generalizes effectively to OOD environments with substantially increased scale and complexity.

\subsection{Craftax Environments}

Finally, we evaluate PACE in Craftax~\cite{matthews2024craftax}, which induces a continually expanding and non-stationary task distribution as new regions, mechanics, and objectives emerge over time. 
This setting tests whether UED methods maintain robust performance under long-horizon exploration and continual changes in task structure.
Following the Craftax-1B Challenge~\cite{matthews2024craftax}, we evaluate agents on the Craftax-Symbolic environment with a budget of 1 billion environment interactions (1B timesteps $\approx$ 6 years of human gameplay.).

We compare PACE with DR and established UED baselines, including PLR, PLR$^{\perp}$, and ACCEL.
Specifically, PLR uses the absolute value of the GAE as evaluation metric. while PLR$^{\perp}$ and ACCEL use the PVL metric. For ACCEL, we adopt the noise mutation operator, where each level $\theta$ is parameterized by the angle vectors of the Perlin noise generator and mutation adds uniform noise to each element, which is the strongest variant reported in prior work.

\begin{figure}[t]
  \centering
  \includegraphics[width=0.85\columnwidth]{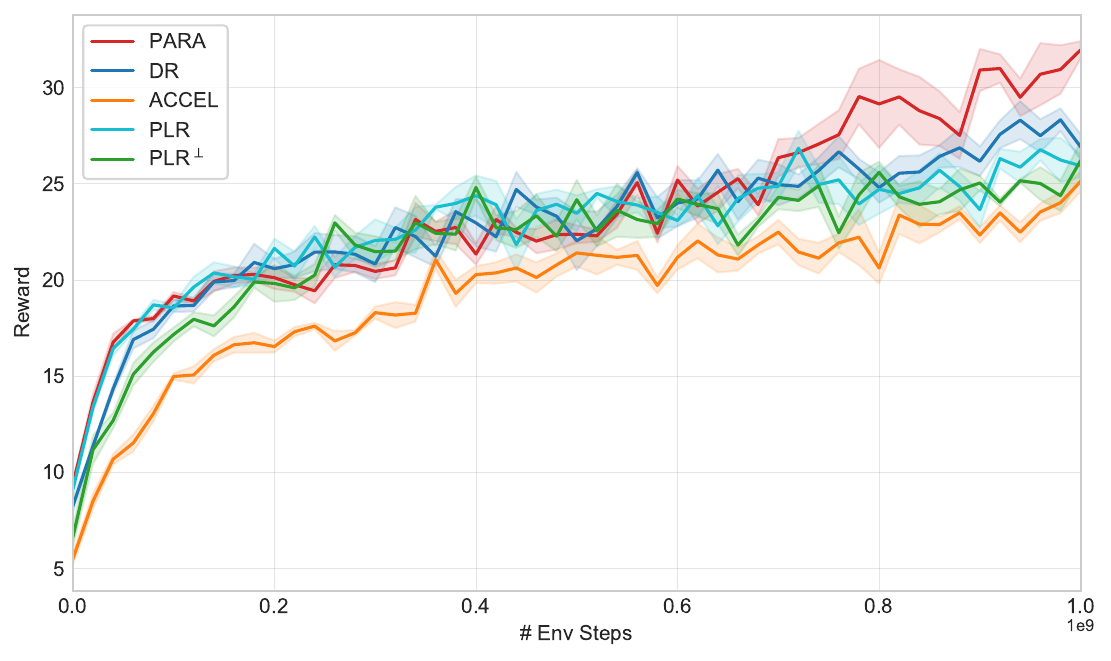}
  \caption{Training reward on Craftax-1B versus environment interactions. Curves show the mean over 10 independent runs and the shaded region denotes 1 standard error.}
  \label{fig:curves}
\end{figure}

We report training reward as a function of environment interactions (Figure~\ref{fig:curves}).
In the early stage of training, all methods achieve similar reward, while differences become more pronounced as training progresses.
As training proceeds, PACE continues to improve steadily in the later phase of training and finishes with the highest reward among all compared methods, indicating more sustained progress under the long-horizon and non-stationary dynamics of Craftax.
As can be seen, the widening gap in the later stage suggests that PACE is better at converting additional experience into progress on harder, later-game outcomes, rather than plateauing early.

\begin{figure}[t]
  \centering
  \includegraphics[width=0.85\columnwidth]{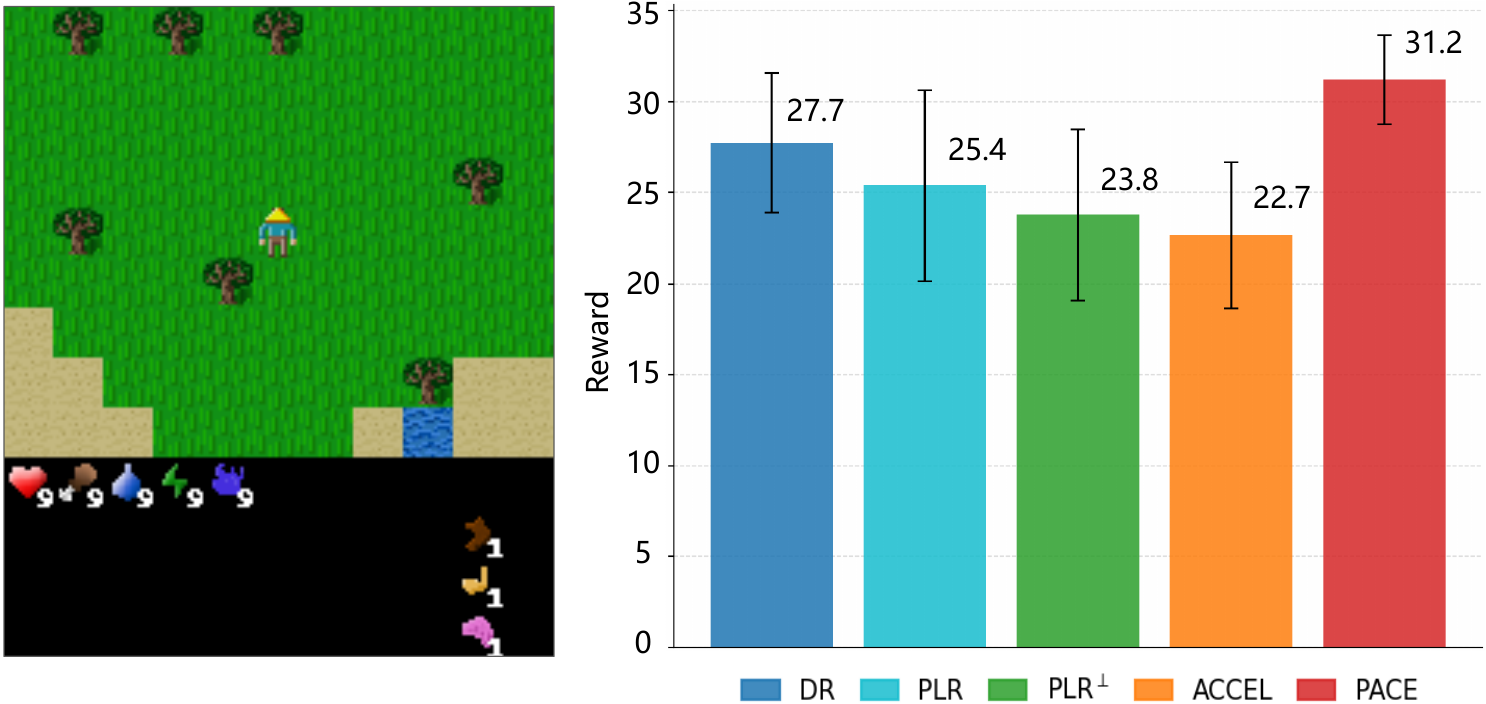}
  \caption{Reward on the Craftax-1B evaluation set after training. Bars show the mean and standard deviation over 10 runs.}
  \label{fig:reward}
\end{figure}

After training for 1 billion timesteps, we evaluate the final saved checkpoints on a fixed set of 20 evaluation levels following the Craftax-1B protocol (Figure~\ref{fig:reward}). 
PACE achieves the highest evaluation reward, outperforming all baselines under the same interaction budget.
We provide per-level results across the 20 evaluation maps in Figure~\ref{fig:craftax_barplot} and Table~\ref{tab:craftax_iqm} in Appendix~\ref{Results} to illustrate that the performance advantage is not driven by a small subset of favorable levels.

\begin{figure}[t]
  \centering
  \includegraphics[width=0.85\columnwidth]{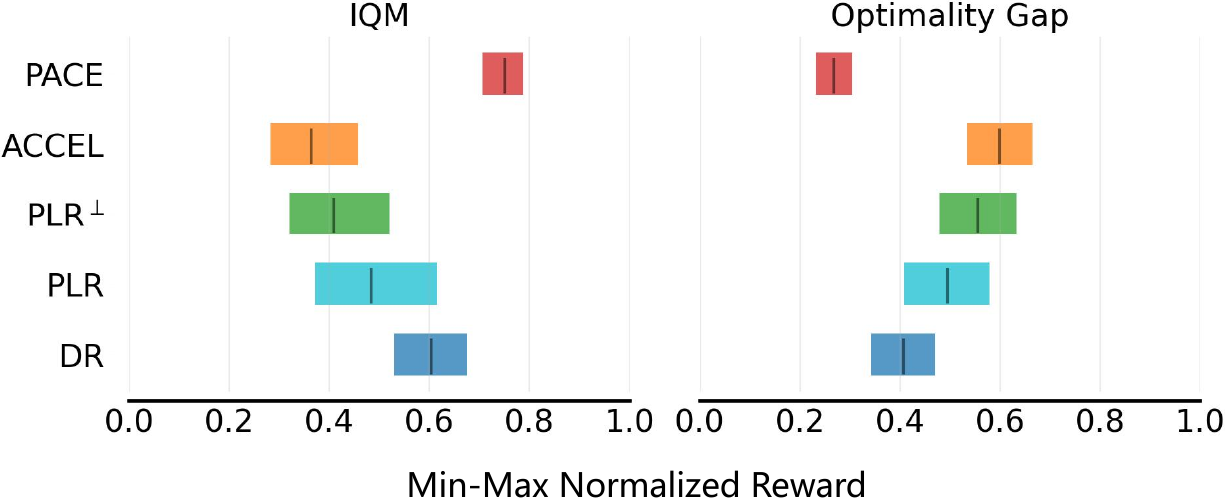}
  \caption{Aggregate OOD test performance on Craftax environments.}
  \label{fig:IQMcra}
\end{figure}

We also apply the same IQM and Optimality Gap metrics to summarize reward distribution across evaluation levels and seeds. Figure~\ref{fig:IQMcra} reports the corresponding results with confidence intervals.
PACE achieves the highest IQM (0.694; CI: [0.532, 0.850]) among all methods, while the strongest baseline under IQM, PLR$^{\perp}$, attains 0.618 (CI: [0.361, 0.832]).
Under the Optimality Gap metric, PACE also yields the smallest gap (0.341; CI: [0.209, 0.473]), improving over all baselines.
Complete numerical results are provided in Table~\ref{tab:craftax_iqm} in Appendix~\ref{Results}.

%% file: chapter/7.Conclusion.tex
\section{Conclusion}

We present Parameter Change Environment Design (PACE), a simple and effective framework for Unsupervised Environment Design that evaluates environments through policy parameter change. Specifically,
PACE uses an intrinsic and low-variance signal that reflects agent realized learning progress, without relying on value estimation, regret approximation, or Monte Carlo return comparison. Theoretically
, A first-order theoretical analysis motivates a direct relationship between environment-induced policy updates and improvement in the policy optimization objective.
This analysis provides a principled foundation for using policy parameter change as an environment value signal. Empirically
results on MiniGrid and large-scale Craftax benchmarks show that PACE consistently improves generalization and remains robust under long-horizon and non-stationary training dynamics.
Overall, PACE offers a practical and principled alternative to existing UED evaluation criteria by grounding environment value in realized learning progress.
More broadly, it reframes environment evaluation around realized optimization progress, and highlights policy parameter change as a reliable signal for guiding environment selection in reinforcement learning.

%% file: chapter/8.appendix.tex
\setcounter{page}{1}
\onecolumn
{
  \centering
  \Large
  \textbf{PACE: Parameter Change for Unsupervised Environment Design}\\
  \vspace{0.5em}Supplementary Material \\
  \vspace{1.0em}
}

\hrule
\vspace{1em}

\section{Appendix: Related Work}
\label{relatedwork}

Unsupervised Environment Design (UED)~\cite{dennis2020emergent} studies how to automatically construct training environment curricula that match an agent current learning stage, with the goal of improving generalization and robustness in unseen environments. Research in this area follows a clear trajectory. Early work emphasizes environment diversity as a heuristic for generalization, later developments introduce game-theoretic or curriculum-based formulations, and recent studies increasingly focus on the alignment between environment evaluation signals and realized learning progress.

\textbf{Early Precursors and Problem Formulation.}
Before the formal introduction of UED, researchers already recognize the importance of environment diversity for generalization. Among early approaches, domain randomization trains policies in large-scale randomized simulators to improve robustness and sim-to-real transfer, and serves as a fundamental baseline for environment diversity~\cite{tobin2017domain}. However, such methods typically sample environment parameters in a static and indiscriminate manner, which limits their ability to adapt to the agent evolving learning dynamics.

In parallel, ideas from open-ended reinforcement learning highlight the potential of environment--agent co-evolution. Representative work such as POET~\cite{wang2019paired} evolves environments and agents jointly and continuously introduces new challenges, demonstrating that environment generation plays a central role in long-horizon learning. Building on this perspective, \cite{dennis2020emergent} provides the first systematic formulation of Unsupervised Environment Design, modeling it as an underspecified POMDP and proposing the PAIRED algorithm. PAIRED casts environment design as an adversarial game, where a teacher maximizes the return gap (regret) between a protagonist and an antagonist to generate structured environments with progressively increasing difficulty. This formulation establishes a clear problem definition and forms the theoretical foundation for subsequent UED methods.

\textbf{Methodological Improvements and Paradigm Shifts.}
Despite its conceptual appeal, PAIRED relies on multi-agent adversarial training, which often introduces instability and substantial computational overhead in practice. These limitations motivate a paradigm shift from explicit adversarial environment generation toward environment selection based on learning signals.
In this context, \cite{jiang2021replay} proposes Prioritized Level Replay (PLR), which becomes a representative approach. PLR maintains a buffer of procedurally generated environments and prioritizes replay according to trajectory-level proxy signals, such as Generalized Advantage Estimation or TD error. Due to its simplicity, stability, and computational efficiency, PLR quickly emerges as a strong baseline in UED research.

Building on PLR, \cite{parker2022evolving} introduces ACCEL, which combines adversarial generation with replay-based selection. By incorporating stabilization techniques such as opponent behavior cloning, ACCEL alleviates some of the instability inherent in adversarial training, reflecting an effort to balance adaptive environment generation with reliable optimization.

\textbf{Theoretical Deepening and Objective Evolution.}
As UED methods are applied to increasingly complex environments and longer training horizons, researchers begin to re-examine the core optimization objectives. Early work largely relies on regret maximization as an environment value signal. However, \cite{beukman2024refining} shows that naive regret formulations may encourage teachers to generate environments that lie beyond the agent current learning capacity, effectively creating learning traps. To address this issue, the authors refine the regret definition and propose the ReMiDi algorithm, which explicitly balances solvability and challenge. This work marks a shift from empirically motivated heuristics toward more principled objective design.

Following this line of inquiry, \cite{monette2025optimisation} develops a unified optimization perspective for UED, analyzing stability and convergence properties across different methods within a common framework. Together, these studies highlight a key trend in UED research from 2024 to 2025: the focus gradually moves away from how to generate environments toward whether environment evaluation signals faithfully reflect actual learning progress.

Motivated by this observation, subsequent work explores alternatives to regret and its approximations. In replay-based methods, environment value is often estimated using proxy metrics such as GAE, TD error, or Monte Carlo returns \cite{jiang2021replay, jiang2021prioritized}. While effective in practice, these metrics provide only indirect measures of learning progress and remain sensitive to value estimation bias, variance amplification, and additional evaluation cost. To further reduce the gap between environment evaluation and the true optimization objective, \cite{li2025marginal} proposes the Marginal Benefit perspective, defining environment value as the performance improvement before and after a policy update on the same environment, instantiated in the MBeDED algorithm. Nevertheless, this approach still relies on extra Monte Carlo evaluations, which introduce nontrivial computational overhead and variance.

\textbf{Multi-Agent Extensions and Emerging Directions.}
In recent years, the UED paradigm extends naturally to multi-agent reinforcement learning settings. In such scenarios, environment design not only controls task difficulty but also systematically shapes interaction structures among agents.
\cite{samvelyan2023maestro} introduces MAESTRO, a representative multi-agent UED approach that automatically generates environments promoting cooperation, competition, and mixed interaction patterns. This work demonstrates that UED can influence emergent collective behavior, rather than merely adjusting difficulty. This direction also connects closely to broader open-ended reinforcement learning efforts, including environment--agent co-evolution frameworks such as POET \cite{wang2019paired}, which emphasize continual expansion of the environment distribution to drive sustained capability growth.

Beyond environment generation, the core idea of UED increasingly generalizes toward \emph{training distribution design}. For example, \cite{ruhdorfer2025unsupervised} proposes Unsupervised Partner Design, which shifts the focus from environment parameters to the automatic construction of diverse training partners with distinct behavioral profiles. By exposing agents to varied interaction partners, this line of work improves zero-shot ad-hoc teamwork and suggests that underspecified factors in UED extend beyond environments themselves.

\textbf{Connections to Automatic Curriculum Learning.}
From a broader perspective, UED can be viewed as a specific instantiation of automatic curriculum learning, where the training distribution adapts dynamically to match the learner's current capabilities \cite{soviany2022curriculum}. Early curriculum learning methods often rely on manually designed task sequences, while later work introduces automated teacher-student frameworks that evolve curricula based on learning progress signals \cite{portelas2020automatic, parker2022evolving}.

UED differs from traditional curriculum learning in a fundamental way. Rather than scheduling or reweighting a fixed set of tasks, UED operates on an underspecified POMDP, where the task distribution itself emerges through automatic design or selection of environment instances \cite{dennis2020emergent}. The curriculum is thus constructed implicitly by instantiating free environment parameters. Recent studies in continual learning further highlight the importance of task ordering and generative task design for robust knowledge accumulation \cite{wang2023progressive, alet2019tailoring}. These works share a core challenge with UED: designing an efficient and stable environment (or task) evaluation signal that is tightly coupled with measurable learning progress, moving beyond simple regret maximization toward more nuanced objectives \cite{beukman2024refining}.

Taken together, prior work establishes UED as a powerful framework for improving generalization through adaptive environment shaping. At the same time, this line of research reveals a persistent tension between theoretical environment value criteria and signals that reliably reflect agent realized learning progress under practical constraints. Resolving this tension remains a central challenge for scalable and robust Unsupervised Environment Design.

\section{Appendix: Complexity and Stability Analysis}
\label{ComplexityAnalysis}

To clarify the efficiency and stability advantages of PACE, we compare it with representative Unsupervised Environment Design (UED) methods.

PAIRED relies on a multi-agent adversarial formulation involving a teacher, an antagonist, and a protagonist. This design introduces substantial computational overhead and often leads to unstable training dynamics. Moreover, its regret-based signal reflects theoretical solvability gaps rather than the agent realized learning progress. Replay-based methods such as PLR and ACCEL use value-based errors, including Generalized Advantage Estimation (GAE), which primarily measure prediction accuracy instead of performance improvement. In contrast, Marginal Benefit (MB) directly targets the objective increment $\Delta J$ and therefore serves as the most direct baseline for comparison.

We analyze PACE and MB from two complementary perspectives: computational complexity and evaluation stability.

\paragraph{Evaluation Complexity.}
MB evaluates environment value through a dual evaluation procedure. After obtaining updated parameters $\theta_{\text{new}}$, it performs $k$ additional rollout episodes of horizon $T$ to estimate the marginal improvement. Let $P$ denote the number of policy parameters. Since each forward pass has complexity $O(P)$, the additional evaluation cost of MB is
\begin{equation}
    C_{\text{MB}} \approx O(k \cdot T \cdot P).
\end{equation}

In contrast, PACE evaluates a level by computing the squared norm of the policy parameter difference. This operation involves only tensor arithmetic in memory and requires no further environment interaction:
\begin{equation}
    C_{\text{PACE}} \approx O(P).
\end{equation}
In typical reinforcement learning settings, $k \cdot T$ ranges from $10^3$ to $10^4$. As a result, PACE reduces the evaluation cost by several orders of magnitude and incurs \emph{no additional sampling cost} beyond the standard training update.

\paragraph{Evaluation Stability.}
The MB estimate $\Delta \hat{J}_{\text{MB}}$ is computed as the difference between two Monte Carlo return estimates. Its variance therefore satisfies
\[
\mathrm{Var}(\Delta \hat{J}_{\text{MB}})
\approx \mathrm{Var}(V_{\text{new}}) + \mathrm{Var}(V_{\text{old}}).
\]
In sparse-reward or highly stochastic environments, this compounded variance can dominate the true improvement signal, which often results in noisy and oscillatory environment evaluation.

By contrast, the PACE score $\|\Delta \theta\|_2^2$ provides a deterministic measurement of the policy parameter update induced by a level. Although the update depends on stochastic training batches, PACE avoids the additional sampling noise introduced by explicit policy re-evaluation. This tight coupling between the evaluation signal and the optimization step yields a more consistent estimate of learning progress, which substantially improves curriculum stability and convergence behavior.



\section{Appendix: Additional Experimental Results}
\label{Results}

\begin{table}[htbp]
\centering
\caption{Total number of environment interactions for a given number of student PPO updates.
All methods use the same student PPO training budget within each environment.
$\dagger$ Craftax uses a batched PPO setup with large-scale parallel environments,
resulting in a substantially larger number of environment interactions per update.}
\label{tab:env_steps}
\small
\begin{tabular}{lcccccc}
\toprule
\textbf{Environment} & \textbf{PPO Updates} & \textbf{PACE} & \textbf{ACCEL} & \textbf{PLR}$^{\perp}$ & \textbf{PLR} & \textbf{DR} \\
\midrule
MiniGrid & $30\,000$ & $245$M & $245$M & $245$M & $245$M & $245$M \\
Craftax & $255^{\dagger}$ & $1.07$B & $1.07$B & $1.07$B & $1.07$B & $1.07$B \\
\bottomrule
\end{tabular}
\end{table}

\subsection{Full Experimental Results}

We report extended evaluation results on a set of 12 held-out MiniGrid levels commonly used in the
UED literature~\cite{parker2022evolving}.
These levels span a range of structural complexities and are designed to test zero-shot
generalization to environments that differ substantially from those encountered during training.
All methods are evaluated after the same number of student PPO updates, following the protocol
used in ACCEL and PLR.

\begin{figure}
\centering
\includegraphics[width=0.85\linewidth]{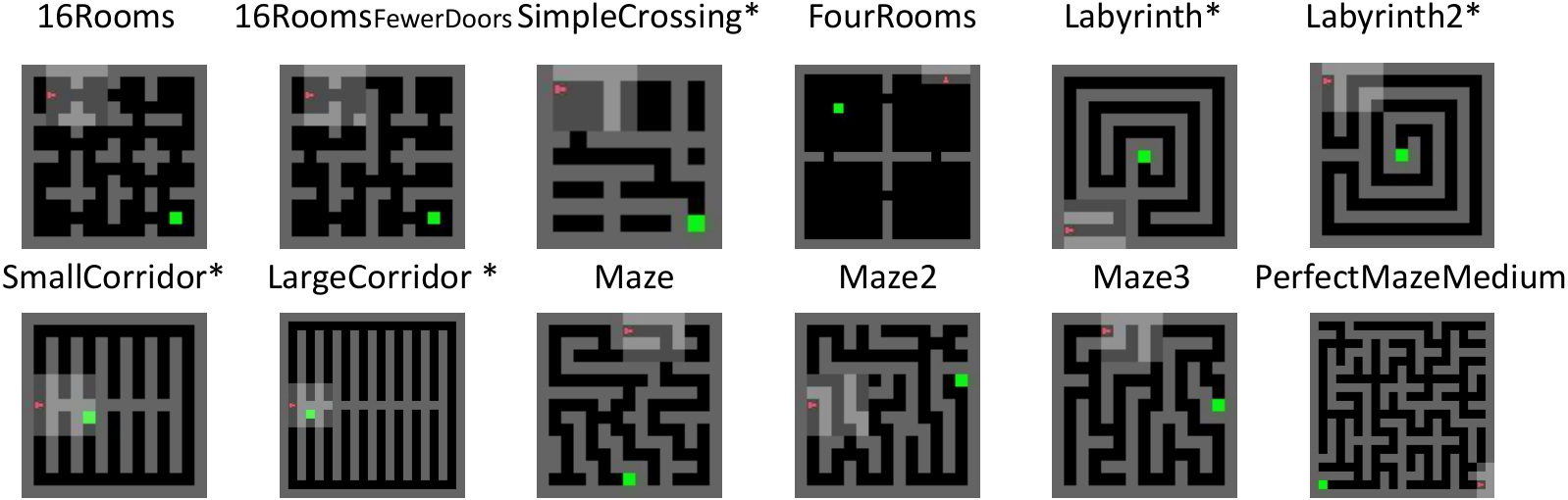}
\caption{MiniGrid zero-shot evaluation levels used for transfer evaluation.}
\label{fig:minigrid_12_levels}
\end{figure}

\begin{figure}
\centering
\includegraphics[width=\linewidth]{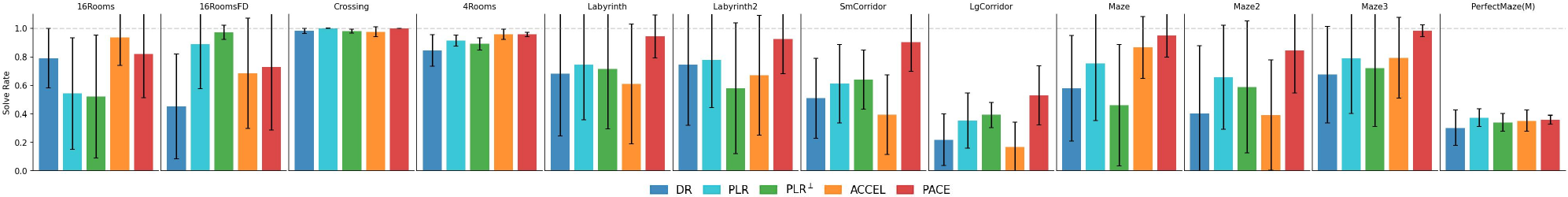}
\caption{Zero-shot transfer results in out-of-distribution mazes. Agents are evaluated for 100 episodes on human-designed mazes.
Plots show mean and standard deviation for each environment, across 10 runs.}
\label{fig:mazesolve}
\end{figure}

\begin{table*}
\centering
\caption{Zero-shot solved rate on 12 held-out MiniGrid levels.
Each entry reports the mean $\pm$ standard deviation over 10 training seeds.
Bold values indicate the highest solved rate for each level.}
\label{tab:full_minigrid_results}
\small
\begin{tabular}{lccccc}
\toprule
\textbf{Level} 
& \textbf{DR} 
& \textbf{PLR} 
& \textbf{PLR}$^{\perp}$ 
& \textbf{ACCEL} 
& \textbf{PACE} \\
\midrule
16Rooms 
& 0.791 $\pm$ 0.209 
& 0.542 $\pm$ 0.391 
& 0.520 $\pm$ 0.431 
& \textbf{0.934} $\pm$ 0.195 
& 0.819 $\pm$ 0.305 \\

16RoomsFD 
& 0.451 $\pm$ 0.368 
& 0.887 $\pm$ 0.310 
& \textbf{0.972} $\pm$ 0.050 
& 0.685 $\pm$ 0.386 
& 0.728 $\pm$ 0.441 \\

Crossing 
& 0.981 $\pm$ 0.018 
& 1.000 $\pm$ 0.001 
& 0.980 $\pm$ 0.014 
& 0.975 $\pm$ 0.035 
& \textbf{1.000} $\pm$ 0.000 \\

4Rooms 
& 0.844 $\pm$ 0.110 
& 0.912 $\pm$ 0.038 
& 0.889 $\pm$ 0.042 
& 0.956 $\pm$ 0.036 
& \textbf{0.957} $\pm$ 0.015 \\

SmCorridor 
& 0.510 $\pm$ 0.281 
& 0.612 $\pm$ 0.274 
& 0.640 $\pm$ 0.207 
& 0.394 $\pm$ 0.280 
& \textbf{0.901} $\pm$ 0.201 \\

LgCorridor 
& 0.218 $\pm$ 0.181 
& 0.354 $\pm$ 0.193 
& 0.393 $\pm$ 0.088 
& 0.168 $\pm$ 0.174 
& \textbf{0.530} $\pm$ 0.207 \\

Labyrinth 
& 0.682 $\pm$ 0.435 
& 0.747 $\pm$ 0.387 
& 0.716 $\pm$ 0.420 
& 0.610 $\pm$ 0.420 
& \textbf{0.943} $\pm$ 0.150 \\

Labyrinth2 
& 0.747 $\pm$ 0.425 
& 0.781 $\pm$ 0.337 
& 0.580 $\pm$ 0.459 
& 0.671 $\pm$ 0.419 
& \textbf{0.923} $\pm$ 0.240 \\

Maze 
& 0.580 $\pm$ 0.370 
& 0.754 $\pm$ 0.399 
& 0.460 $\pm$ 0.425 
& 0.865 $\pm$ 0.216 
& \textbf{0.950} $\pm$ 0.152 \\

Maze2 
& 0.403 $\pm$ 0.473 
& 0.658 $\pm$ 0.364 
& 0.588 $\pm$ 0.463 
& 0.392 $\pm$ 0.387 
& \textbf{0.844} $\pm$ 0.299 \\

Maze3 
& 0.675 $\pm$ 0.338 
& 0.791 $\pm$ 0.389 
& 0.721 $\pm$ 0.408 
& 0.794 $\pm$ 0.283 
& \textbf{0.983} $\pm$ 0.042 \\

PerfectMaze(M) 
& 0.302 $\pm$ 0.124 
& \textbf{0.373} $\pm$ 0.061 
& 0.339 $\pm$ 0.061 
& 0.352 $\pm$ 0.074 
& 0.359 $\pm$ 0.030 \\

\midrule
\textbf{Mean} 
& 0.599 $\pm$ 0.230 
& 0.701 $\pm$ 0.202
& 0.650 $\pm$ 0.214
& 0.650 $\pm$ 0.270 
& \textbf{0.828} $\pm$ \textbf{0.198} \\
\bottomrule
\end{tabular}
\end{table*}

\begin{table}
\centering
\caption{Aggregate performance across 12 held-out MiniGrid levels.
We report the Interquartile Mean (IQM) and Optimality Gap of solved rates,
computed using the \texttt{rliable} library.
Values in brackets denote 95\% confidence intervals.
Bold values indicate the best performance for each metric.}
\label{tab:iqm_gap_results}
\small
\begin{tabular}{lcc}
\toprule
\textbf{Method} 
& \textbf{IQM} $\uparrow$ 
& \textbf{Optimality Gap} $\downarrow$ \\
\midrule
DR 
& 0.657 [0.444, 0.842] 
& 0.401 [0.280, 0.521] \\

PLR 
& 0.808 [0.607, 0.950] 
& 0.299 [0.200, 0.407] \\

PLR$^{\perp}$ 
& 0.733 [0.521, 0.908] 
& 0.350 [0.239, 0.458] \\

ACCEL 
& 0.739 [0.475, 0.937] 
& 0.350 [0.206, 0.498] \\

PACE 
& \textbf{0.964} [0.800, 0.997] 
& \textbf{0.172} [0.077, 0.278] \\
\bottomrule
\end{tabular}
\end{table}

\subsection{Craftax Evaluation Results}

We report evaluation results on the Craftax benchmark following the Craftax-1B protocol.
After training for a fixed interaction budget, we evaluate the final policy checkpoints
on a held-out set of 20 Craftax levels.
Unlike MiniGrid, performance in Craftax is measured by episodic reward rather than solved rate.
All reported results therefore use reward-based metrics, consistent with the evaluation
setup described in the main paper.

\begin{figure*}
\centering
\includegraphics[width=\textwidth]{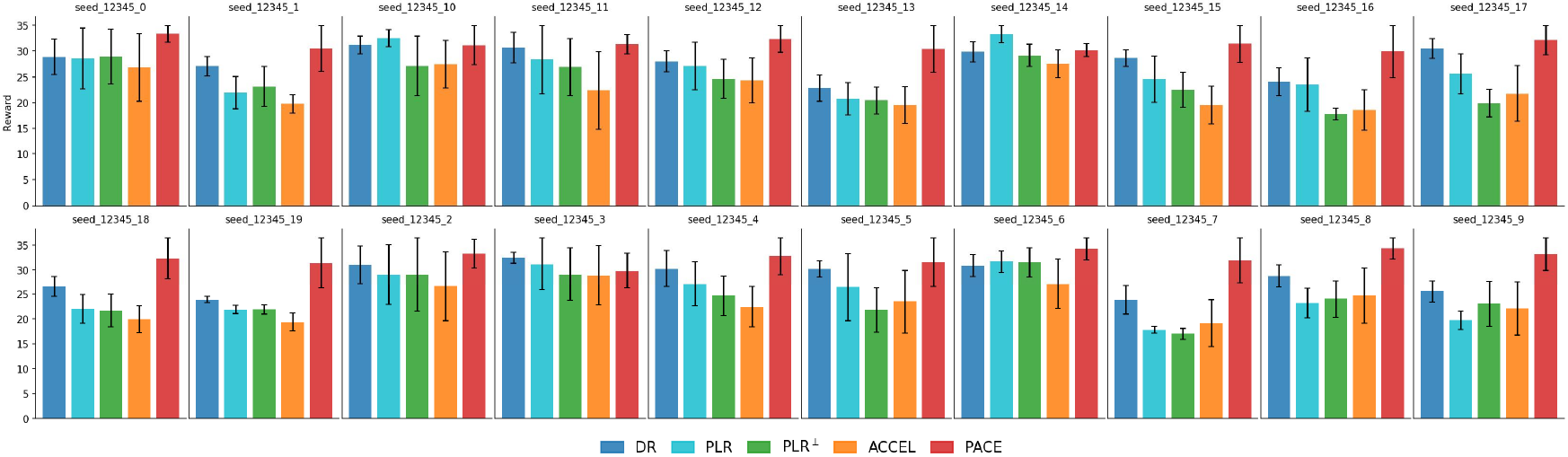}
\caption{Evaluation reward on 20 held-out Craftax levels.
Bars report the mean episodic reward and error bars indicate one standard deviation
across training seeds.
All methods are evaluated using the same final policy checkpoints.
}
\label{fig:craftax_barplot}
\end{figure*}

\begin{table*}
\centering
\caption{Per-level evaluation reward on 20 Craftax levels.
Each entry reports the mean $\pm$ standard deviation of episodic reward
over 10 training seeds.
The Mean row reports the average and standard deviation across levels.
}
\label{tab:craftax_per_level}
\small
\begin{tabular}{lccccc}
\toprule
\textbf{Level}
& \textbf{DR}
& \textbf{PLR}
& \textbf{PLR}$^{\perp}$
& \textbf{ACCEL}
& \textbf{PACE} \\
\midrule
0  & 28.871 $\pm$ 3.433 & 28.545 $\pm$ 5.869 & 28.950 $\pm$ 5.355 & 26.828 $\pm$ 6.521 & \textbf{33.315} $\pm$ 1.629 \\
1  & 24.893 $\pm$ 1.718 & 20.184 $\pm$ 2.909 & 21.277 $\pm$ 3.524 & 18.193 $\pm$ 1.643 & \textbf{28.072} $\pm$ 4.066 \\
2  & 30.349 $\pm$ 3.676 & 28.403 $\pm$ 5.957 & 28.402 $\pm$ 7.290 & 26.073 $\pm$ 6.845 & \textbf{32.559} $\pm$ 2.824 \\
3  & 32.687 $\pm$ 1.120 & \textbf{31.375} $\pm$ 5.362 & 29.265 $\pm$ 5.382 & 29.050 $\pm$ 6.128 & 30.016 $\pm$ 3.608 \\
4  & 25.841 $\pm$ 3.165 & 23.198 $\pm$ 3.848 & 21.143 $\pm$ 3.427 & 19.205 $\pm$ 3.488 & \textbf{28.012} $\pm$ 3.201 \\
5  & 26.968 $\pm$ 1.472 & 23.602 $\pm$ 6.079 & 19.450 $\pm$ 4.023 & 21.009 $\pm$ 5.712 & \textbf{28.111} $\pm$ 4.446 \\
6  & 31.158 $\pm$ 2.214 & 32.004 $\pm$ 2.210 & 31.798 $\pm$ 2.922 & 27.393 $\pm$ 5.102 & \textbf{34.553} $\pm$ 2.275 \\
7  & 22.352 $\pm$ 2.704 & 16.676 $\pm$ 0.631 & 15.938 $\pm$ 1.034 & 17.936 $\pm$ 4.445 & \textbf{29.909} $\pm$ 4.267 \\
8  & 28.412 $\pm$ 2.234 & 22.949 $\pm$ 2.936 & 23.777 $\pm$ 3.729 & 24.449 $\pm$ 5.563 & \textbf{33.932} $\pm$ 2.117 \\
9  & 24.446 $\pm$ 2.094 & 18.862 $\pm$ 1.704 & 22.103 $\pm$ 4.395 & 21.192 $\pm$ 5.198 & \textbf{31.729} $\pm$ 3.149 \\
10 & 31.349 $\pm$ 1.725 & \textbf{32.661} $\pm$ 1.679 & 27.255 $\pm$ 5.785 & 27.566 $\pm$ 4.607 & 31.293 $\pm$ 3.809 \\
11 & 33.722 $\pm$ 3.215 & 31.144 $\pm$ 7.251 & 29.566 $\pm$ 6.015 & 24.571 $\pm$ 8.218 & \textbf{34.439} $\pm$ 2.051 \\
12 & 27.643 $\pm$ 1.979 & 26.768 $\pm$ 4.525 & 24.309 $\pm$ 3.709 & 23.997 $\pm$ 4.269 & \textbf{31.926} $\pm$ 2.560 \\
13 & 23.464 $\pm$ 2.597 & 21.359 $\pm$ 3.222 & 21.012 $\pm$ 2.678 & 20.068 $\pm$ 3.707 & \textbf{31.244} $\pm$ 4.665 \\
14 & 30.899 $\pm$ 2.039 & \textbf{34.453} $\pm$ 1.724 & 30.154 $\pm$ 2.231 & 28.534 $\pm$ 2.783 & 31.233 $\pm$ 1.296 \\
15 & 29.230 $\pm$ 1.643 & 25.090 $\pm$ 4.537 & 22.961 $\pm$ 3.464 & 19.952 $\pm$ 3.718 & \textbf{32.044} $\pm$ 3.608 \\
16 & 24.151 $\pm$ 2.652 & 23.628 $\pm$ 5.150 & 17.880 $\pm$ 1.063 & 18.694 $\pm$ 3.910 & \textbf{30.034} $\pm$ 5.059 \\
17 & 32.087 $\pm$ 2.053 & 26.955 $\pm$ 4.072 & 20.935 $\pm$ 2.790 & 22.915 $\pm$ 5.686 & \textbf{33.784} $\pm$ 2.969 \\
18 & 26.609 $\pm$ 2.081 & 22.049 $\pm$ 2.909 & 21.706 $\pm$ 3.282 & 19.940 $\pm$ 2.698 & \textbf{32.380} $\pm$ 4.131 \\
19 & 19.323 $\pm$ 0.525 & 17.701 $\pm$ 0.658 & 17.715 $\pm$ 0.762 & 15.664 $\pm$ 1.438 & \textbf{25.368} $\pm$ 4.137 \\
\midrule
\textbf{Mean}
& 27.723 $\pm$ 3.839
& 25.380 $\pm$ 5.233
& 23.780 $\pm$ 4.688
& 22.661 $\pm$ 4.005
& \textbf{31.198 $\pm$ 2.439} \\
\bottomrule
\end{tabular}
\end{table*}

\begin{table}
\centering
\caption{Aggregate performance across 20 Craftax evaluation levels.
We report the Interquartile Mean (IQM) and Optimality Gap of episodic reward,
computed using the \texttt{rliable} library.
Values in brackets denote 95\% confidence intervals.
Bold values indicate the best performance for each metric.}
\label{tab:craftax_iqm}
\small
\begin{tabular}{lcc}
\toprule
\textbf{Method}
& \textbf{IQM} $\uparrow$
& \textbf{Optimality Gap} $\downarrow$ \\
\midrule
DR
& 0.603 [0.530, 0.675]
& 0.407 [0.343, 0.470] \\

PLR
& 0.484 [0.372, 0.615]
& 0.495 [0.409, 0.580] \\

PLR$^{\perp}$
& 0.409 [0.320, 0.520]
& 0.556 [0.479, 0.633] \\

ACCEL
& 0.364 [0.283, 0.458]
& 0.599 [0.534, 0.664] \\

PACE
& \textbf{0.750} [0.706, 0.788]
& \textbf{0.268} [0.232, 0.305] \\
\bottomrule
\end{tabular}
\end{table}

\subsection{Hyperparameters}

The majority of hyperparameters are inherited from prior work on unsupervised environment design~\cite{dennis2020emergent,jiang2021prioritized,jiang2021replay,parker2022evolving}, with only minor modifications.
For both MiniGrid and Craftax, we conduct a grid search over the level replay buffer size
$\{500, 1000, 2000, 4000, 6000, 8000\}$, the replay probability
$\{0.3, 0.5, 0.6, 0.7, 0.8\}$, and the temperature parameter
$\beta \in \{0.3, 0.5, 0.6, 0.7, 0.8, 0.9, 1.0\}$.
Hyperparameters are selected based on performance on held-out validation levels.

For MiniGrid, we follow the evaluation protocol used in ACCEL and PLR, and adopt the same set of
12 human-designed levels as validation environments.
For Craftax, we follow the experimental setup from~\cite{matthews2024craftax},
and evaluate on the Craftax-1B Challenge using the Craftax-Symbolic environment.
We randomly generate 20 evaluation levels and use them as validation levels for hyperparameter selection.

The final hyperparameters used in all experiments are reported in Table~\ref{tab:hyperparameters}.

\begin{table}[t]
\centering
\caption{Hyperparameters used in all experiments.}
\label{tab:hyperparameters}
\small
\begin{tabular}{lcc}
\toprule
\textbf{Parameter} & \textbf{MiniGrid} & \textbf{Craftax} \\
\midrule
\multicolumn{3}{l}{\textbf{PPO}} \\
Learning rate & $1 \times 10^{-4}$ & $3 \times 10^{-4}$ ($2 \times 10^{-4}$ DR) \\
Max gradient norm & $0.5$ & $1.0$ \\
Number of PPO updates & $30{,}000$ & $255$ \\
Rollout length & $256$ & -- \\
Inner rollout length & -- & $64$ \\
Outer rollout length & -- & $64$ \\
Number of environments & $32$ & $1024$ \\
Minibatches per update & $1$ & $2$ \\
Discount factor $\gamma$ & $0.995$ & $0.995$ \\
PPO epochs & $5$ & $5$ \\
Clip range & $0.2$ & $0.2$ \\
GAE parameter $\lambda$ & $0.98$ & $0.90$ \\
Entropy coefficient & $1 \times 10^{-3}$ & $1 \times 10^{-2}$ \\
Value loss coefficient & $0.5$ & $0.5$ \\
\midrule
\multicolumn{3}{l}{\textbf{PACE}} \\
Scoring function & Parameter change & Parameter change \\
Level buffer size $K$ & $1000$ & $1000$ \\
Replay probability $p$ & $0.7$ & $0.7$ \\
Staleness coefficient & $0.6$ & $0.6$ \\
Temperature $\beta$ & $0.8$ & $0.8$ \\
Minimum fill ratio $\rho$ & $0.5$ & $0.5$ \\
Prioritization & rank & rank \\
Exploratory gradient updates & False & True \\
Number of walls & $100$(random) & -- \\
\midrule
\multicolumn{3}{l}{\textbf{ACCEL}} \\
Scoring function & MaxMC & PVL \\
Number of edits & $5$ & $100$(Noise) \\
Exploratory gradient updates & False & False \\
Number of walls & $0$ & -- \\
\midrule
\multicolumn{3}{l}{\textbf{PLR}$^{\perp}$} \\
Scoring function & MaxMC & PVL \\
Exploratory gradient updates & False & False \\
Number of walls & $100$(random) & -- \\
\midrule
\multicolumn{3}{l}{\textbf{PLR}} \\
Scoring function & $\lvert \text{GAE} \rvert$ & $\lvert \text{GAE} \rvert$ \\
Exploratory gradient updates & True & True \\
Number of walls & $100$(random) & -- \\
\bottomrule
\end{tabular}
\end{table}

%% file: example_paper.bib
@article{shakya2023reinforcement,
  title={Reinforcement learning algorithms: A brief survey},
  author={Shakya, Ashish Kumar and Pillai, Gopinatha and Chakrabarty, Sohom},
  journal={Expert Systems with Applications},
  volume={231},
  pages={120495},
  year={2023},
  publisher={Elsevier}
}

@article{vinyals2019grandmaster,
  title={Grandmaster level in StarCraft II using multi-agent reinforcement learning},
  author={Vinyals, Oriol and Babuschkin, Igor and Czarnecki, Wojciech M and Mathieu, Micha{\"e}l and Dudzik, Andrew and Chung, Junyoung and Choi, David H and Powell, Richard and Ewalds, Timo and Georgiev, Petko and others},
  journal={nature},
  volume={575},
  number={7782},
  pages={350--354},
  year={2019},
  publisher={Nature Publishing Group}
}

@article{jain2024recent,
  title={Recent developments of game theory and reinforcement learning approaches: A systematic review},
  author={Jain, Garima and Kumar, Arun and Bhat, Shahid Ahmad},
  journal={IEEE Access},
  volume={12},
  pages={9999--10011},
  year={2024},
  publisher={IEEE}
}

@article{elguea2023review,
  title={A review on reinforcement learning for contact-rich robotic manipulation tasks},
  author={Elguea-Aguinaco, {\'I}{\~n}igo and Serrano-Mu{\~n}oz, Antonio and Chrysostomou, Dimitrios and Inziarte-Hidalgo, Ibai and B{\o}gh, Simon and Arana-Arexolaleiba, Nestor},
  journal={Robotics and Computer-Integrated Manufacturing},
  volume={81},
  pages={102517},
  year={2023},
  publisher={Elsevier}
}

@article{tang2025deep,
  title={Deep reinforcement learning for robotics: A survey of real-world successes},
  author={Tang, Chen and Abbatematteo, Ben and Hu, Jiaheng and Chandra, Rohan and Mart{\'\i}n-Mart{\'\i}n, Roberto and Stone, Peter},
  journal={Annual Review of Control, Robotics, and Autonomous Systems},
  volume={8},
  number={1},
  pages={153--188},
  year={2025},
  publisher={Annual Reviews}
}

@article{towers2024gymnasium,
  title={Gymnasium: A standard interface for reinforcement learning environments},
  author={Towers, Mark and Kwiatkowski, Ariel and Terry, Jordan and Balis, John U and De Cola, Gianluca and Deleu, Tristan and Goul{\~a}o, Manuel and Kallinteris, Andreas and Krimmel, Markus and KG, Arjun and others},
  journal={arXiv preprint arXiv:2407.17032},
  year={2024}
}

@article{andrychowicz2020learning,
  title={Learning dexterous in-hand manipulation},
  author={Andrychowicz, OpenAI: Marcin and Baker, Bowen and Chociej, Maciek and Jozefowicz, Rafal and McGrew, Bob and Pachocki, Jakub and Petron, Arthur and Plappert, Matthias and Powell, Glenn and Ray, Alex and others},
  journal={The International Journal of Robotics Research},
  volume={39},
  number={1},
  pages={3--20},
  year={2020},
  publisher={SAGE Publications Sage UK: London, England}
}

@article{dennis2020emergent,
  title={Emergent complexity and zero-shot transfer via unsupervised environment design},
  author={Dennis, Michael and Jaques, Natasha and Vinitsky, Eugene and Bayen, Alexandre and Russell, Stuart and Critch, Andrew and Levine, Sergey},
  journal={Advances in neural information processing systems},
  volume={33},
  pages={13049--13061},
  year={2020}
}

@article{french1999catastrophic,
  title={Catastrophic forgetting in connectionist networks},
  author={French, Robert M},
  journal={Trends in cognitive sciences},
  volume={3},
  number={4},
  pages={128--135},
  year={1999},
  publisher={Elsevier}
}

@inproceedings{jiang2021prioritized,
  title={Prioritized level replay},
  author={Jiang, Minqi and Grefenstette, Edward and Rockt{\"a}schel, Tim},
  booktitle={International Conference on Machine Learning},
  pages={4940--4950},
  year={2021},
  organization={PMLR}
}

@article{jiang2021replay,
  title={Replay-guided adversarial environment design},
  author={Jiang, Minqi and Dennis, Michael and Parker-Holder, Jack and Foerster, Jakob and Grefenstette, Edward and Rockt{\"a}schel, Tim},
  journal={Advances in Neural Information Processing Systems},
  volume={34},
  pages={1884--1897},
  year={2021}
}

@inproceedings{parker2022evolving,
  title={Evolving curricula with regret-based environment design},
  author={Parker-Holder, Jack and Jiang, Minqi and Dennis, Michael and Samvelyan, Mikayel and Foerster, Jakob and Grefenstette, Edward and Rockt{\"a}schel, Tim},
  booktitle={International Conference on Machine Learning},
  pages={17473--17498},
  year={2022},
  organization={PMLR}
}

@inproceedings{li2025marginal,
  title={Marginal Benefit Driven RL Teacher for Unsupervised Environment Design},
  author={Li, Dexun and Li, Wenjun and Varakantham, Pradeep},
  booktitle={Proceedings of the AAAI Conference on Artificial Intelligence},
  volume={39},
  number={17},
  pages={18253--18261},
  year={2025}
}

@inproceedings{tobin2017domain,
  title={Domain randomization for transferring deep neural networks from simulation to the real world},
  author={Tobin, Josh and Fong, Rachel and Ray, Alex and Schneider, Jonas and Zaremba, Wojciech and Abbeel, Pieter},
  booktitle={2017 IEEE/RSJ international conference on intelligent robots and systems (IROS)},
  pages={23--30},
  year={2017},
  organization={IEEE}
}

@inproceedings{pinto2017supervision,
  title={Supervision via competition: Robot adversaries for learning tasks},
  author={Pinto, Lerrel and Davidson, James and Gupta, Abhinav},
  booktitle={2017 IEEE International Conference on Robotics and Automation (ICRA)},
  pages={1601--1608},
  year={2017},
  organization={IEEE}
}

@article{schulman2017proximal,
  title={Proximal policy optimization algorithms},
  author={Schulman, John and Wolski, Filip and Dhariwal, Prafulla and Radford, Alec and Klimov, Oleg},
  journal={arXiv preprint arXiv:1707.06347},
  year={2017}
}

@article{agarwal2021deep,
  title={Deep reinforcement learning at the edge of the statistical precipice},
  author={Agarwal, Rishabh and Schwarzer, Max and Castro, Pablo Samuel and Courville, Aaron C and Bellemare, Marc},
  journal={Advances in neural information processing systems},
  volume={34},
  pages={29304--29320},
  year={2021}
}

@article{matthews2024craftax,
  title={Craftax: A lightning-fast benchmark for open-ended reinforcement learning},
  author={Matthews, Michael and Beukman, Michael and Ellis, Benjamin and Samvelyan, Mikayel and Jackson, Matthew and Coward, Samuel and Foerster, Jakob},
  journal={arXiv preprint arXiv:2402.16801},
  year={2024}
}

@article{coward2024jaxued,
  title={JaxUED: A simple and useable UED library in Jax},
  author={Coward, Samuel and Beukman, Michael and Foerster, Jakob},
  journal={arXiv preprint arXiv:2403.13091},
  year={2024}
}

@article{chevalier2018minimalistic,
  title={Minimalistic gridworld environment for openai gym (2018)},
  author={Chevalier-Boisvert, Maxime and Willems, Lucas and Pal, Suman},
  journal={URL https://github. com/maximecb/gym-minigrid},
  volume={6},
  year={2018}
}

@article{hafner2021benchmarking,
  title={Benchmarking the spectrum of agent capabilities},
  author={Hafner, Danijar},
  journal={arXiv preprint arXiv:2109.06780},
  year={2021}
}

@article{kuttler2020nethack,
  title={The nethack learning environment},
  author={K{\"u}ttler, Heinrich and Nardelli, Nantas and Miller, Alexander and Raileanu, Roberta and Selvatici, Marco and Grefenstette, Edward and Rockt{\"a}schel, Tim},
  journal={Advances in Neural Information Processing Systems},
  volume={33},
  pages={7671--7684},
  year={2020}
}

@article{wang2019paired,
  title={Paired open-ended trailblazer (poet): Endlessly generating increasingly complex and diverse learning environments and their solutions},
  author={Wang, Rui and Lehman, Joel and Clune, Jeff and Stanley, Kenneth O},
  journal={arXiv preprint arXiv:1901.01753},
  year={2019}
}

@article{beukman2024refining,
  title={Refining minimax regret for unsupervised environment design},
  author={Beukman, Michael and Coward, Samuel and Matthews, Michael and Fellows, Mattie and Jiang, Minqi and Dennis, Michael and Foerster, Jakob},
  journal={arXiv preprint arXiv:2402.12284},
  year={2024}
}

@article{monette2025optimisation,
  title={An optimisation framework for unsupervised environment design},
  author={Monette, Nathan and Letcher, Alistair and Beukman, Michael and Jackson, Matthew T and Rutherford, Alexander and Goldie, Alexander D and Foerster, Jakob N},
  journal={arXiv preprint arXiv:2505.20659},
  year={2025}
}

@article{samvelyan2023maestro,
  title={MAESTRO: Open-ended environment design for multi-agent reinforcement learning},
  author={Samvelyan, Mikayel and Khan, Akbir and Dennis, Michael and Jiang, Minqi and Parker-Holder, Jack and Foerster, Jakob and Raileanu, Roberta and Rockt{\"a}schel, Tim},
  journal={arXiv preprint arXiv:2303.03376},
  year={2023}
}

@article{ruhdorfer2025unsupervised,
  title={Unsupervised Partner Design Enables Robust Ad-hoc Teamwork},
  author={Ruhdorfer, Constantin and Bortoletto, Matteo and Oei, Victor and Penzkofer, Anna and Bulling, Andreas},
  journal={arXiv preprint arXiv:2508.06336},
  year={2025}
}

@article{soviany2022curriculum,
  title={Curriculum learning: A survey},
  author={Soviany, Petru and Ionescu, Radu Tudor and Rota, Paolo and Sebe, Nicu},
  journal={International Journal of Computer Vision},
  volume={130},
  pages={1526--1565},
  year={2022}
}

@article{portelas2020automatic,
  title={Automatic curriculum learning for deep RL: A short survey},
  author={Portelas, R{\'e}my and Colas, C{\'e}dric and Hofmann, Katja and Oudeyer, Pierre-Yves},
  journal={arXiv preprint arXiv:2003.04664},
  year={2020}
}

@inproceedings{alet2019tailoring,
  title={Tailoring: Encoding inductive biases by generating student tasks},
  author={Alet, Ferran and Schneider, Martin F and Lozano-Perez, Tomas and Kaelbling, Leslie Pack},
  booktitle={ICML Workshop on Automated Machine Learning},
  year={2019}
}

@article{wang2023progressive,
  title={Progressive continual learning for spoken keyword spotting},
  author={Wang, Hsin-Wei and Huang, Chao-Han Huck and Shi, Jiahong and Chung, Yu-An and others},
  journal={IEEE Journal of Selected Topics in Signal Processing},
  year={2023}
}
